\documentclass[times,twocolumn,final]{elsarticle}

\usepackage[colorinlistoftodos]{todonotes}
\usepackage{medima}
\usepackage{framed,multirow}

\usepackage{amssymb}
\usepackage{latexsym}

\usepackage{url}
\usepackage{xcolor}
\usepackage{subcaption}
\usepackage{amsmath}
\usepackage{algorithm}
\usepackage{algpseudocode}
\usepackage{pifont}
\usepackage{booktabs}

\usepackage{amsmath,amsfonts,bm}

\def\eqref#1{equation~\ref{#1}}
\def\1{\bm{1}}

\def\vDelta{{\bm{\Delta}}}
\def\vtheta{{\bm{\theta}}}

\def\vt{{\bm{t}}}

\def\vv{{\bm{v}}}

\def\vx{{\bm{x}}}

\def\mW{{\bm{W}}}

\def\mWtheta{\mW_\vtheta}
\DeclareMathAlphabet{\mathsfit}{\encodingdefault}{\sfdefault}{m}{sl}
\SetMathAlphabet{\mathsfit}{bold}{\encodingdefault}{\sfdefault}{bx}{n}

\newcommand{\R}{\mathbb{R}}

\definecolor{newcolor}{rgb}{.8,.349,.1}

\journal{Medical Image Analysis}

\begin{document}

\verso{Riccardo Taiello \textit{et~al.}}

\begin{frontmatter}
	
	\title{Privacy Preserving Image Registration}
	
	\author[1,2,3]{Riccardo \snm{Taiello}\corref{cor1}}
	\cortext[cor1]{Corresponding author:  \href{mailto:riccardo.taiello@inria.fr}{riccardo.taiello@inria.fr}}
	\author[2]{Melek \snm{Önen}}
	%\fntext[fn1]{This is author footnote for second author.}
        \author[2]{ Francesco \snm{Capano}}
        \author[3]{ Olivier \snm{Humbert}}
	\author[1,3]{Marco \snm{Lorenzi}}
	
	\address[1]{Epione Research Group, Inria, Sophia Antipolis, France}
	\address[2]{EURECOM, France}
	\address[3]{Université Côte d’Azur, France}

% 	\received{1 May 2013}
% 	\finalform{10 May 2013}
% 	\accepted{13 May 2013}
% 	\availableonline{15 May 2013}
% 	\communicated{S. Sarkar}

	\begin{abstract}
		%%%
		Image registration is a key task in medical imaging applications, allowing to represent medical images in a common spatial reference frame. Current approaches to image registration are generally based on the assumption that the content of the images is usually accessible in clear form, from which the spatial transformation is subsequently estimated. This common assumption may not be met in practical applications, since the sensitive nature of medical images may ultimately require their analysis under privacy constraints, preventing to openly share the image content.
		In this work, we formulate the problem of image registration under a privacy preserving regime, where images are assumed to be confidential and cannot be disclosed in clear. 
		We derive our privacy preserving image registration framework by extending classical registration paradigms to account for advanced cryptographic tools, such as secure multi-party computation and homomorphic encryption, that enable the execution of operations without leaking the underlying data. To overcome the problem of performance and scalability of cryptographic tools in high dimensions, we propose several techniques to optimize the image registration operations by using gradient approximations, and by revisiting the use of homomorphic encryption trough packing, to allow the efficient encryption and multiplication of large matrices.
        We focus on registration methods of increasing complexity, including rigid,  affine, and non-linear registration based on cubic splines or diffeomorphisms parametrized by time-varying velocity fields. In all these settings, we demonstrate how the registration problem can be naturally adapted for accounting to privacy-preserving operations, and illustrate the effectiveness of PPIR on a variety of registration tasks.
 		%We demonstrate our privacy preserving framework in linear and non-linear registration problems, evaluating its accuracy and scalability with respect to standard, non-private counterparts. Our results show that privacy preserving image registration is feasible and can be adopted in sensitive medical imaging applications.
		%%%%
	\end{abstract}
	
	
\end{frontmatter}

%\linenumbers

\section{Introduction}
%Image Registration is a crucial task in medical imaging applications, it aims at spatially transforming a moving image to match a given template image.
Image Registration is a crucial task in medical imaging applications, allowing to spatially align imaging features between two or multiple scans. Registration methods are today a central component of state-of-the-art methods for atlas-based segmentation \citep{shattuck2009online,cardoso2013steps}, morphological and functional analysis \citep{dale1999cortical,ashburner2000voxel}, multi-modal data integration \citep{heinrich2011non}, and longitudinal analysis \citep{reuter2010highly,ashburner2013symmetric}.
Typical registration paradigms are based on a given transformation model (e.g. affine or non-linear), a cost function and an associated optimization routine. A large number of image registration approaches have been proposed in the literature over the last decades, covering a variety of assumptions on the spatial transformations, cost functions, image dimensionality and optimization strategy \citep{schnabel2016advances}.
Image registration is the workhorse of many real-life medical imaging software and applications, including public web-based services for automated segmentation and labelling of medical images. Using these services generally requires uploading and exchanging medical images over the Internet, to subsequently perform image registration with respect to one or multiple (potentially proprietary) atlases. Besides these classical medical imaging use-cases, emerging paradigms for collaborative data analysis, such as Federated Learning (FL) \citep{mcmahan2017communication}, have been proposed to enable analysis of medical images in multicentric scenarios for performing group analysis \citep{gazula2021decentralized} and distributed machine learning \citep{kaissis2021end, zerka2020blockchain}. %, as well as methods for encrypting medical images to secure their transmission over networks \citep{hua2018medical}. 
However, in these settings, typical medical imaging tasks such as spatial alignment and downstream operations are generally not possible without disclosing the image information.

Due to the evolving juridical landscape on data protection, medical image analysis tools need to be adapted to guarantee compliance with regulations currently existing in many countries, 
such as the European General Data Protection Regulation (GDPR) \footnote{\url{https://gdpr-info.eu/}}, or the US Health Insurance Portability and Accountability Act (HIPAA)\footnote{\url{https://www.hhs.gov/hipaa/index.html}}. Medical imaging information falls within the realm of personal health data \citep{lotan2020medical} and its sensitive nature should ultimately require the analysis under privacy preserving constraints, for instance by preventing to share the image content in clear form. 

Advanced cryptographic tools  hold great potential in sensitive data analysis problems (e.g., \citep{Lauter21eprint}).
Examples of such approaches are Secure-Multi-Party-Computation (MPC)  \citep{yao1982protocols} and Homomorphic Encryption (HE) \citep{rivest1978data}.
While MPC allows multiple parties to jointly compute a common function over their private inputs and discover no more than the output of this function, HE enables computation on encrypted data without disclosing either the input data or the result of the computation.

This work presents \emph{privacy-preserving image registration} (PPIR), a new methodological framework allowing image registration under privacy constraints. To this end, we reformulate the image registration problem to integrate cryptographic tools, namely MPC or FHE, thus preserving the privacy of the image data. Due to the well-known scalability issues of such cryptographic techniques, we investigate strategies for the practical use of PPIR. % through gradient approximations, array packing and matrix partitioning.
In our experiments, we evaluate the effectiveness of PPIR on a variety of registration tasks and medical imaging modalities. Our results demonstrate the feasibility of PPIR and pave the way for the application of secured image registration in  sensitive  medical  imaging applications.
\section{Background}
\begin{algorithm}[!t]
	\begin{algorithmic}[1]

		\small
		\algnewcommand\algorithmicinput{\textbf{Input:}}
		\algnewcommand\Input{\item[\algorithmicinput]}
		\algnewcommand\algorithmicoutput{\textbf{Output:}}
		\algnewcommand\Output{\item[\algorithmicoutput]}
		\Input 
		\Statex $\triangleright$ Moving image $I$ 
		\Statex $\triangleright$ Template image $J$ 
		\Statex $\triangleright$ Distance function $f$ 
		\Statex $\triangleright$ Spatial transformation $\mWtheta$, parameterized by $\vtheta$ 
		\Statex $\triangleright$ convergence threshold $\epsilon$ 

		\Output
		\Statex $\triangleright$ Transformed image $I(\mWtheta)$ after convergence is reached 
		\Statex
		\Function{ImageRegistration}{$I ,J, \mWtheta$}:
		\State $\vtheta \longleftarrow $ {\sc InitializeParameters()} 
        \Repeat 

		\State $e \longleftarrow$ $f \big[I(\mWtheta), J\big]$

		\State $G \longleftarrow $ {$\frac{\partial f}{\partial \vtheta}$}
		\State $H \longleftarrow $ {$\frac{\partial^2 f}{\partial \vtheta^2}$}
        
        \State $\vDelta\vtheta \longleftarrow  H^{-1} \cdot G$
		
		  \State $\vtheta \longleftarrow \vtheta + \vDelta\vtheta$
		
		 \Until{$\| \vDelta \vtheta\| \leq \epsilon$}
		 \Statex
		\State \textbf{return} $I(\mWtheta)$, $e$
		\EndFunction
	\end{algorithmic}
    \caption{IR via Gauss-Newton optimization}\label{fig:algotirthm}
\end{algorithm}

\begin{figure*}
\centering
\begin{subfigure}{.5\textwidth}
  \centering
  \includegraphics[width=.65\linewidth]{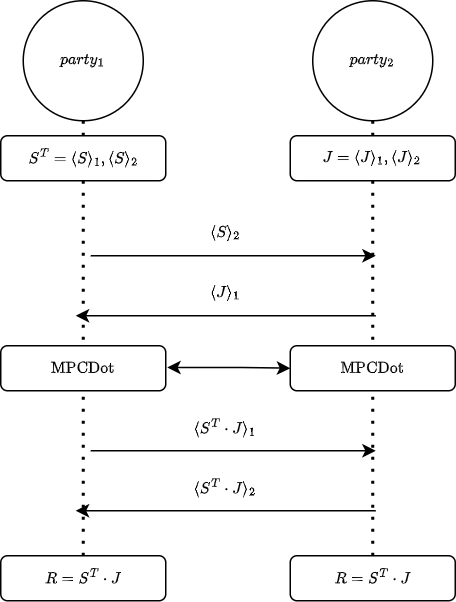}
  \caption{PPIR(MPC)}
  \label{fig:mpc}
\end{subfigure}%
\begin{subfigure}{.5\textwidth}
  \centering
  \includegraphics[width=.67\linewidth]{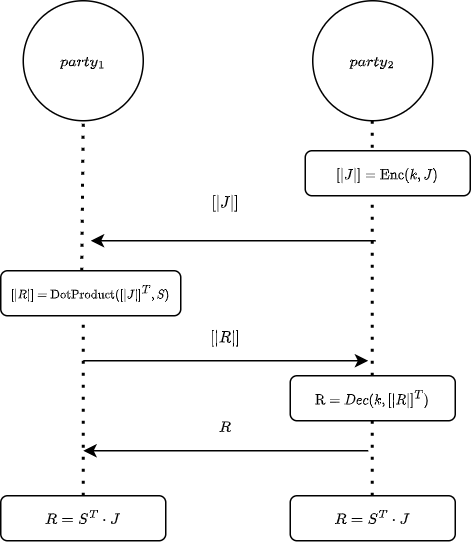}
  \caption{PPIR(FHE)-v1}
  \label{fig:he}
\end{subfigure}
\caption{Optimization of SSD loss: proposed framework to compute matrix-vector multiplication $S^T \cdot J$ based on PPIR(MPC) and PPIR(FHE)-v1.}
\label{fig:framework-ssd}
\end{figure*}

Given images $I, J: \mathbb{R}^d \mapsto \mathbb{R}$, image registration (IR) aims at estimating the parameters $\vtheta$ of a spatial transformation $\mWtheta\in \mathbb{R}^d \mapsto \mathbb{R}^d$, either linear or non-linear,  maximizing the spatial overlap between $J$ and the transformed image $I(\mWtheta)$, by minimizing a registration loss function $f$:
\begin{equation}
    \vtheta^{*}= \text{argmin}_{\vtheta}~ 
    f \left(I(\mWtheta(\vx)), J(\vx)\right).
    \label{eq: objective}
\end{equation}
The loss $f$ can be any similarity measure, e.g., the Sum of Squared Differences (SSD), the negative Mutual Information (MI), or normalized cross correlation (CC).
Equation (\ref{eq: objective}) can be typically optimized through gradient-based methods, where the parameters $\vtheta$ are iteratively updated until convergence. %(line 10 of Algorithm 1).  
In particular, when using a Gauss-Newton optimization scheme (Algorithm \ref{fig:algotirthm}), the update of the spatial transformation can be computed through Equation (\ref{eq:delta}): 
  \begin{equation}
        \label{eq:delta}
    \vDelta\vtheta = H^{-1} \cdot G,
 \end{equation}
where $G = \frac{\partial f}{\partial \vtheta}$ is the Jacobian and  $H = \frac{\partial^2 f}{\partial \vtheta^2}$ the Hessian of $f$.
\textcolor{black}{Besides the Gauss-Newton schemes proposed in the field of IR \citep{pennec1999understanding, 2009-FAIR}, gradient-based techniques are classically adopted to solve the IR task, for example in diffeomorphic image registration problems \citep{ashburner2007fast, avants2011reproducible}.}

In all these cases we consider a scenario with two parties, $party_1$, and $party_2$, whereby $party_1$ owns image $I$ and $party_2$ owns image $J$. The parties wish to collaboratively optimize the image registration problem without disclosing their respective images to each other. We assume that only $party_1$ has access to the transformation parameters $\vtheta$ and that it is also responsible for computing the update at each optimization step.  \textcolor{black}{In what follows, we introduce the basic notation to develop PPIR based on different registration frameworks. We focus on registration methods of increasing complexity, including \emph{(i)} rigid, \emph{(ii)} affine, and \emph{(iii) }non-linear registration based on cubic splines or diffeomorphisms parametrized by time-varying velocity fields (large deformation diffeomorphic metric mapping, LDDMM) \citep{beg2005computing}. In all these settings, we demonstrate how the registration problem can be naturally adapted for accounting to privacy-preserving operations, and illustrate the effectiveness of PPIR on a variety of registration tasks.}

\section{Analysis of classical IR loss functions under a privacy-preserving perspective}
\label{sec:IR}
In this section we review typical loss functions used in image registration, and analyze the related requirements for privacy-preserving optimization. 
\subsection{Optimization of SSD loss}
\label{sec:ssd}
A typical loss function to be optimized during the registration process is the sum of squared intensity differences (SSD) evaluated on the set of image coordinates:
\begin{equation}
    \textsc{SSD}(I,J, \vtheta )= \text{argmin}_{\vtheta} \sum_{\vx} [I(\mWtheta(\vx)) - J(\vx)]^{2}
    \label{eq: objective_ssd}
\end{equation}
with Jacobian: 
\begin{equation}
    G = \sum_{\vx}  S(\vx) \cdot (I(\mWtheta(\vx)) - J(\vx)),
    \label{eq:jacobian_ssd}
 \end{equation}
where the quantity \begin{equation}
    S(\vx) =\nabla I(\vx)\frac{\partial\mWtheta(\vx)}{\partial\vtheta}
\end{equation} quantifies image and transformation gradients, and \begin{equation}
    H = \sum_{\vx} \left(\nabla I(\vx)\frac{\partial\mWtheta(\vx)}{\partial\vtheta}\right)^T\left(\nabla I(\vx)\frac{\partial\mWtheta(\vx)}{\partial\vtheta}\right)
    \label{eq:hessian_ssd}
\end{equation} is the second order term obtained from Equation (\ref{eq: objective_ssd}) through linearization \citep{pennec1999understanding, baker2004lucas}. The solution to this problem requires the joint availability of both images ${I}$ and ${J}$, as well as of the gradients of $I$ and of $\mWtheta$. In a privacy-preserving setting, this information cannot be disclosed, and the computation of Equation (\ref{eq:delta}) is therefore impossible.
We note that to calculate the update of the registration $\vDelta\vtheta$ of Equation (\ref{eq:delta}), the only operation that requires the joint availability of information from both parties is the term 
$R = \sum_{\vx} S(\vx) \cdot J(\vx)$, which can be computed a matrix-vector multiplication of vectorized quantities $ R = S^T\cdot J$.
\subsection{Mutual Information}
Mutual Information quantifies the joint information content between the intensity distributions of the two images.
This is calculated from the joint probability distribution function (PDF):
\begin{equation}
    MI(I,J, \vtheta) = 
    \text{argmin}_{\vtheta} - \sum_{r, t}p(r, t;\vtheta)\log\left(\frac{p(r, t;\vtheta)}{p(r; \vtheta)p(t)}\right)
    \label{eq:objective_mi}
\end{equation}
where, given $N_r$ and $N_t$ the maximum intensity  for respectively $I$ and  $J$, we define $r \in [0;N_r-1] \subseteq \mathbb{N}$ and $t \in [0;N_t-1] \subseteq \mathbb{N}$ as the range of discretized intensity values of $I$ and $J$, respectively.
A Parzen window \citep{parzen1961mathematical} is used to generate continuous estimates of the underlying intensity distributions, thereby reducing the effects of quantization from interpolation, and discretization from binning the data.
Let $\psi_I^3 : \mathbb{R} \mapsto \left[0,1\right]$ be a cubic spline Parzen window, and let $\psi_J^0 : \mathbb{R} \mapsto \left[0,1\right]$ be a zero-order spline Parzen window.
The smoothed joint histogram of $I$ and $J$ \citep{viola1997alignment,mattes2003pet} is given by: 
    \begin{align*}
    &p(r, t;\vtheta) =\\ &{\frac{1}{N_{\vx}}}\sum_{\vx}\psi_I^3 \left(r - \frac{I(\mWtheta(\vx)) - I(\mWtheta(\vx))^{\circ}}{\mathbf{\Delta}b_r}\right) \cdot \psi_J^0 \left(t - \frac{J(\vx)  - J^{\circ}}{\mathbf{\Delta}b_t} \right)
    \end{align*}

In the above formula, the intensity values of $I$ and $J$ are normalized by their respective minimum (denoted by $I(\mWtheta)^{\circ}$ and $J^{\circ}$), and by the bin size (respectively  $\mathbf{\Delta}b_r$ and $\mathbf{\Delta}b_t$), to fit into the specified number of bins ($b_r$ or $b_t$) of the intensity distribution. The final value for $p(r,t;\vtheta)$ is computed by normalizing by $N_{\vx}$, the number of sampled voxels.
Marginal probabilities are simply obtained by summing along one axis of the PDF, that is, $p(r)=\sum_t p(r,t;\vtheta)$ and $p(t)=\sum_r p(r,t;\vtheta)$.
Let the matrices $A_I^3 \in \mathbb{R}^{N_{\vx} \times N_{r}}$ and $B_J^0 \in \mathbb{R}^{N_{\vx} \times N_{t}}$ be defined as:
\begin{align*}
    A_I^3(\vx, r; \vtheta) = \psi_I^3 \left(r - \frac{I(\mWtheta(\vx)) - I(\mWtheta)^{\circ}}{\mathbf{\Delta}b_r}\right) 
\end{align*}
and
\begin{align*}
    B_J^0(\vx, t) = \psi_J^0\left(t - \frac{J(\vx)  - J^{\circ}}{\mathbf{\Delta}b_t} \right),
\end{align*}
the discretized joint PDF can be rewritten in a matrix form via the multiplication:
\begin{equation}
    P = {\frac{1}{N_{\vx}}} \cdot (A_I^3)^T \cdot B_J^0.
    \label{eq:joint_pdf}
\end{equation}

The first derivative of the joint PDF is calculated as follows \citep{dowson2006unifying}:
\begin{align*}
    &\frac{\partial p(r,t;\vtheta)}{\partial \vtheta}
    = - \frac{1}{N_{\vx}} \cdot \sum_{\vx} B_J^0(\vx,t) \cdot \frac{\partial \psi_I^3(\epsilon)}{\partial \epsilon} \cdot \frac{\partial \epsilon}{\partial I(\mWtheta(\vx))} \cdot \frac{\partial I(\mWtheta(\vx))}{\partial \vtheta}
\end{align*}
where $\epsilon = \epsilon(\vx,r;\vtheta) = r -\frac{I(\mWtheta(\vx)) - I(\mWtheta)^{\circ}}{\mathbf{\Delta}b_r}$ is the input of the cubic spline.
We also introduce the tensor $C_I^3 \in \mathbb{R}^{N_{\vx} \times N_{r} \times |\vtheta|}$ defined as $C_I^3(\vx,r;\vtheta) = \frac{\partial \psi_I^3(\epsilon)}{\partial \epsilon} \cdot \frac{\partial \epsilon}{\partial I(\mWtheta(\vx))} \cdot \frac{\partial I(\mWtheta(\vx))}{\partial \vtheta}$, to write the discretized first derivative as:
\begin{equation}
    P'= - \frac{1}{N_{\vx}} \cdot (B_J^0)^T \cdot C_I^3,
    \label{eq:joint_der_pdf}
\end{equation}
\textcolor{black}{where $P' \in \mathbb{R}^{N_{t} \times N_{r} \times |\vtheta|}$.}

The Jacobian of the MI is obtained from the
chain rule and takes the form:
\textcolor{black}{\begin{align*}
    G = P'\log\left(\frac{P}{P_I}\right),
\end{align*}}
while the linearized Hessian \citep{dowson2007mutual} can be written as:
\textcolor{black}{\begin{align*}
    H = P'^{T}P'\left(\frac{1}{P} - \frac{1}{P_I}\right),
\end{align*}
where $P_I = \sum_t{p}(r,t, \vtheta)$ is a vector which defines the discretized marginal PDF of the moving image.}

The derivatives can be easily calculated from the properties of
B-splines since we have $\frac{\partial \psi_I^3}{\partial \epsilon} = \psi^2_I(\epsilon + \frac{1}{2}) - \psi^2_I(\epsilon - \frac{1}{2})$.
In a privacy-preserving scenario, to calculate the update of the registration $\vDelta\vtheta$ of Equation (\ref{eq:objective_mi}), two operations require the joint availability of information from both parties, which are the matrix 
$P$ of Equation (\ref{eq:joint_pdf}) and the matrix $P'$ of Equation (\ref{eq:joint_der_pdf}).

\textcolor{black}{\subsection{ Cross Correlation with  Advanced Normalization Tools}
In the Advanced Normalization Tools (ANTs) introduced by \citep{avants2008symmetric}, the normalized cross-correlation (CC) loss was specified in the context of diffeomorphic image registration. Let $\phi$ define a diffeomorphism over the domain $\Omega = \R^d$, parameterized by a time-varying velocity field $\vv(\vx,t)$. In the ANTs setting, inverse consistency is obtained by optimizing the $CC$ loss with respect to both forward and backward (inverse) transformations, here denoted by $\phi_1(\vx, t)$ and $\phi_2(\vx, t)$, and parameterized by velocity fields $\vv_1 (\vx, t)$ and $\vv_2 (\vx, t)$ respectively. In particular, both images $I$ and $J$ are simultaneously warped towards a ``half-way'' space, to obtain $I_1 = I(\phi_1(\vx, 0.5))$ and $J_2 = J(\phi_2(\vx, 0.5))$.
The CC loss is thus defined as:
\color{black}
\[
CC(\vx) = \frac{\sum_{\vx_i}(\bar{I}(\vx_i), \bar{J}(\vx_i))^2}{\sum_{\vx_i}(\bar{I}(\vx_i))^2\sum_{\vx_i}(\bar{J}(\vx_i))^2} = \frac{D^2}{E F}
\]
%let first define $I_1 = I(\phi_1(\vx, 0.5))$ , $J_2 = J(\phi_2(\vx, 0.5))$,
\color{black}
where $\bar{I}(\vx_i) = \left(I_1(\vx_i) - \mu_{I_1}(\vx)\right)$ and $\bar{J}(\vx_i) = \left(J_2(\vx_i) - \mu_{J_2}(\vx)\right)$ quantify the images appearance at location $\vx_i$, with respect to the average intensity $\mu_{I_1}$ and $\mu_{J_2}$ measuread in a local window of size $M$.
%, where $\vx$ is at the center of a $m^d$ window, $\mu$ is the mean value within the window centered at $\vx$ and $\vx_i$ iterates trough that window.
Coherently with the LDDMM formulation, the variational optimization problem is defined as:
\begin{align*}
E_{CC} &= \inf_{\phi_1} \inf_{\phi_2} \int_{t=0}^{0.5} \left\| \vv_1 (\vx, t) \right\|^2_L + \left\| \vv_2(\vx, t) \right\|^2_L \, \text{d}t \\
&\quad+ \int_{\Omega} CC(\vx) \, \text{d}\Omega.
\end{align*}
where $L$ is a linear operator prescribing a norm on the velocity fields acting as a regularizer.
The equation for the derivative of the forward update is given by:
\begin{align}
\label{eq:fw}
\begin{aligned}
    \nabla_{\phi_1(\vx, 0.5)} CC(\vx) &= 2L\vv_1(\vx, 0.5) + \frac{2D}{EF} \\
    &\quad \times \left( \bar{J}(\vx) - \frac{D}{E} \bar{I}(\vx) \right) |D \phi_1| \nabla \bar{I}(\vx),
\end{aligned}
\end{align}
while the derivative of the backward update is analogously given by:
\begin{align}
\label{eq:bw}
\begin{aligned}
     \nabla_{\phi_2(\vx, 0.5)} CC(\vx) &= 2L\vv_2(\vx, 0.5) + \frac{2D}{EF} \\
     &\quad \times \left( \bar{I}(\vx) - \frac{D}{F} \bar{J}(\vx) \right) |D \phi_2| \nabla \bar{J}(\vx)
\end{aligned}
\end{align}
% where $bw(\vx) = (\bar{I}(\vx) - \frac{D}{F} \bar{J}(\vx) )$ and its discretized version is $BW = (\bar{I} - \frac{D}{F} \bar{J} )$
In privacy-preserving scenario, the sensitive terms carrying private image information are $\frac{2D}{EF}$, $(\bar{J}(\vx) - \frac{D}{E} \bar{I}(\vx) )$ and $(\bar{I}(\vx) - \frac{D}{F} \bar{J}(\vx))$, which must therefore be computed in a privacy-preserving regime.
}

\section{Building blocks for Secure Computation}\label{sec:secure_computation}
After introducing in Section 1 the IR optimization problem and the related functionals addressed in this work, in this section, we review the standard privacy-preserving techniques that will be employed to develop PPIR.

\subsection{Secure Multi-Party Computation}
Introduced by \citep{yao1982protocols},  MPC is a cryptographic tool that allows multiple parties to jointly compute a common function over their private inputs (secrets) and discover no more than the output of this function. Among existing MPC protocols, additive secret sharing consists of first splitting every secret  $s$ into additive shares $\langle s \rangle_i$, such that $\sum_{i=1}^{n} \langle s\rangle_i = s$, where  $n$ is the number of collaborating parties. Each party $i$ receives one share $\langle s \rangle_i$ and executes an arithmetic circuit in order to obtain the final output of the function.
In this paper, we adopt the two-party computation protocol defined in SPDZ \citep{cryptoeprint:2011:535}, whereby the actual function is mapped into an arithmetic circuit and all computations are performed within a finite ring with modulus $Q$. %Each operation is translated into its MPC version. 
Additions consist of locally adding shares of secrets, while multiplications require interaction between parties. Following \citep{cryptoeprint:2011:535}, SPDZ defines: \textcolor{black}{\textsc{MPCMul}
to compute element-wise multiplication, \textsc{MPCDot} to compute matrix-vector multiplication and \textsc{MPCMatMul} to compute matrix-matrix multiplication.} These operations are performed within an honest but curious protocol. 
\subsection{Homomorphic Encryption}
Initially introduced by Rivest et al. in \citep{rivest1978data}, Homomorphic Encryption (HE) enables the execution of operations over encrypted data without disclosing either the input data or the result of the computation. 
Hence, $party_1$ encrypts the input with its public key and sends the encrypted input to $party_2$. In turn, $party_2$ evaluates a circuit over this encrypted input and sends the result, which still remains encrypted, back to $party_1$ which can finally decrypt them. Among various HE schemes,  CKKS \citep{cheon2017homomorphic} supports the execution of all operations on encrypted real values and is considered a levelled homomorphic encryption (LHE) scheme. The supported operations are: \textsc{Sum} (+), Element-wise multiplication ($\ast$) and \textsc{DotProduct}.
With CKKS, an input vector is mapped to a polynomial and further encrypted with a public key in order to obtain a pair of polynomials $c=(c_0, c_1)$. The original function is further mapped into a set of operations that are supported by CKKS, which are executed over $c$. 
The performance and security of CKKS depend on multiple parameters including the degree of the polynomial $N$, which is usually sufficiently large (e.g. $N=4096$, or $N=8192$).

\section{Methods: from IR to PPIR}\label{sec:methods}

% We present now PPIR, in which the cryptographic tools of Section 2 are integrated in the registration frameworks introduced in Section 1. To this end, in the following sections we investigate separately the secure optimization of the metric SSD and MI.
In this section, we describe the privacy preserving variants of the three IR methods described in Section \ref{sec:IR}. We propose two versions of PPIR for SSD according to the underlying cryptographic tool, namely PPIR(MPC), integrating MPC, and PPIR(FHE), integrating FHE. In the case of MI and NCC, we focus on the design of the MPC-variant, due to the non-negligible computational overhead of FHE in these applications. Finally, we also  study rigid point cloud registration and describe its privacy-preserving variant in Appendix \ref{appendix:pointcloud}.
%As previously mentioned, the operations of SSD, MI, and, NCC are first revisited in order to be compatible with cryptographic tools. To this end, in the following sections, we first integrate the secure optimization of these three cost functions separately and then propose improvements to the protocols to enhance scalability when applied to large-dimensional imaging information.

\subsection{PPIR based on SSD}
As mentioned in Section \ref{sec:ssd}, when optimizing the SSD cost, the only sensitive operation that must be jointly executed by the parties is the matrix-vector multiplication: $ R = S^T\cdot J$, where $S^T$ is only known to $party_1$ and $J$ to $party_2$.
Figure \ref{fig:framework-ssd} illustrates how cryptographic tools are employed to ensure privacy during registration.  

With MPC (Figure \ref{fig:mpc}), $party_1$ secretly shares the matrix $S^T$ to obtain $(\langle S \rangle_1, \langle S \rangle_2)$, while $party_2$ secretly shares the image $J$ to obtain $(\langle J \rangle_1, \langle J \rangle_2)$. Each party also receives its corresponding share, so that $party_1$ holds ($\langle S \rangle_1, \langle J \rangle_1)$ and $party_2$ holds $(\langle S \rangle_2, \langle J \rangle_2)$. The parties execute a circuit with \textsc{MPCMul} operations to calculate the 2-party dot product between $S^T$ and $J$. The parties further synchronize to allow $party_1$ to obtain the product and finally to calculate $\vDelta \vtheta$ (Equations (\ref{eq:jacobian_ssd}) and (\ref{eq:hessian_ssd})).

When using FHE (Figure \ref{fig:he}), $party_2$ first uses a FHE key $k$ to encrypt ${J}$ and obtains $\llbracket {J} \rrbracket \leftarrow \textsc{Enc}(k, {J})$. This encrypted image is sent to $party_1$, who computes the encrypted matrix-vector multiplication $\llbracket R \rrbracket$. In this framework, only vector $J$ is encrypted, and therefore $party_1$ executes scalar multiplications and additions in the encrypted domain only (which are less costly than multiplications over two encrypted inputs). The encrypted result $\llbracket R \rrbracket$ is sent back to $party_2$, which can obtain the result by decryption: $R = \textsc{Dec}(k, \llbracket R \rrbracket)$. Finally, $party_1$ receives $R$ in clear form and can therefore compute $\vDelta \vtheta$.

Thanks to the privacy and security guarantees of these cryptographic tools, during the entire registration procedure, the content of the image data $S$ and $J$ is never disclosed to the opposite party.
% However, 

\subsection{PPIR based on MI}

% \begin{subfigure}{.5\textwidth}
%   \centering
%   \includegraphics[width=.67\linewidth]{joint_pdf_fhe.drawio.png}
%   \caption{PPIR(FHE)}
%   \label{fig:joint_pdf_fhe}
% \end{subfigure}

% \begin{figure}[!htp]
%     \centering
%     \includegraphics[width=.5\linewidth]{cc.drawio.png}
%     \caption{\textcolor{black}{Proposed framework to calculate $\frac{2D}{EF}$ based on PPIR(MPC).}}
%     \label{fig:joint_deriv_cc}
% \end{figure}

% \begin{figure}[!htp]
%     \centering
%     \includegraphics[width=.5\linewidth]{cc_fw.drawio.png}
%     \caption{\textcolor{black}{Proposed framework to calculate $(\bar{I} - \frac{D}{F} \bar{J})$ based on PPIR(MPC).}}
%    \label{fig:fw_cc}
% \end{figure}
In the MPC variant to calculate the joint PDF $P$ in a privacy-preserving manner, the quantity $A_I^3$ is only known to $party_1$, while the quantity $B_J^0$ to $party_2$ (Section \ref{sec:IR}.2). With reference to Supplementary Figure \ref{fig:joint_pdf_mpc}, $party_1$ secretly shares matrix $(A^3_I)^T = (\langle A_I^3 \rangle_1, \langle A_I^3 \rangle_2)$, while $party_2$ does the same with $B_J^0 = (\langle B_J^0 \rangle_1, \langle B_J^0 \rangle_2)$.
Each party also receives its corresponding share: $party_1$ now holds $(\langle A_I^3 \rangle_1, \langle B_J^0 \rangle_1)$, and $party_2$ holds $(\langle B_J^0 \rangle_2, \langle A_I^3 \rangle_2)$. The parties execute a circuit with \textsc{MPCMatMul} operation to calculate the 2-party matrix multiplication between $(A_I^3)^T$ and $B_J^0$.
The next operation carried out in the privacy-preserving setting is the computation of the first derivative of Equation (\ref{eq:joint_der_pdf}). Supplementary Figure \ref{fig:joint_deriv_pdf_mpc} illustrates the MPC variant, where $party_1$ only knows $C^3_I$ and $party_2$ only knows $B^0_J$.
Initially, $party_1$ secretly shares the matrix $C^3_I = (\langle C^3_I \rangle_1, \langle C^3_I \rangle_2)$, while $party_2$ does the same with $(B^0_J)^T = (\langle B^0_J \rangle_1, \langle B^0_J \rangle_2)$. 
Each party also receives its corresponding share, namely: $party_1$ holds $(\langle C^3_I \rangle_1, \langle B^0_J \rangle_1)$ and $party_2$ holds $(\langle C^3_I \rangle_2, \langle B^0_J \rangle_2)$. The parties also execute a circuit with the \textsc{MPCMatMul} between $(B^0_J)^T$ and $C^3_I$. 
\begin{table*}[]
\color{black}
\resizebox{1.0\textwidth}{!}{
\begin{tabular}{llllll}
\toprule
Dataset              & Dimension                                  & Modality & Registration Type    & Loss function & PETs                                 \\
\cmidrule{1-6}
2D  Point Cloud      & 193 points                                 & Mono     & Rigid                & SSD           & MPC and FHE-v1                       \\
2D Whole body PET    & 1260 × 1090 voxels                         & Mono     & Affine/Cubic splines & SSD           & MPC+URS/GMS, FHE-v1 and v2 + URS/GMS \\
2D Brain MRI         & 121 × 121 voxels                           & Mono     & Cubic splines        & SSD           & MPC, FHE-v1 and v2                   \\
3D Brain MRI and PET & 180 × 256 × 256 and 160 × 160 × 96  voxels & Multi    & Affine               & MI            & MPC                                  \\
3D Abdomen MR and CT & 192 × 160 × 192 voxels                     & Multi    & Diffeomorphic (ANTs) & CC            & MPC                            \\
\bottomrule
\end{tabular}
}
\caption{\textcolor{black}{Overview of the datasets used in the study. PETs: Privacy Enhancing Technologies.}}
\label{table:dataset}
\end{table*}

\textcolor{black}{\subsection{PPIR based on ANTS CC loss} According to Section \ref{sec:IR}.3, $party_1$ has access to $\Bar{I}$ and $E$, whereas $party_2$ has access to $\Bar{J}$ and $F$. The computation begins with the optimization of the CC term, specifically the quantity $\frac{2D}{EF}$.  In the initial phase, as illustrated in Supplementary Figure \ref{fig:joint_deriv_cc}, $party_1$ and  $party_2$ secretly share $\Bar{I} = (\langle \Bar{I}\rangle_1, \langle \Bar{I} \rangle_2)$, and $\Bar{J} = (\langle \Bar{J}\rangle_1, \langle \Bar{J} \rangle_2)$, respectively. The subsequent multiplication of these shared values results in the computation of the shares of $D = (\langle D\rangle_1, \langle D \rangle_2)$, which is never reconstructed.}
\textcolor{black}{
In the next step of the protocol, $party_1$ secretly shares $\frac{1}{E}$, and $party_2$ secretly shares $\frac{1}{F}$. Through a multiplication of the shares of $D$, $\frac{1}{E}$, and $\frac{1}{F}$, both parties collectively obtain the final value of $\frac{2D}{EF}$.}

\textcolor{black}{The other two terms that need to be jointly computed are the third term of Equation (\ref{eq:fw}) and (\ref{eq:bw}), namely $(\bar{J} - \frac{D}{E} \bar{I} )$ and $(\bar{I} - \frac{D}{F} \bar{J})$, reported in Supplementary Figure \ref{fig:fw_cc}. In the case of $(\bar{J} - \frac{D}{E} \bar{I} )$, $party_1$ secretly shares $\Bar{I}$ and $\frac{1}{E}$, while $party_2$ secretly shares $\Bar{J}$. Since both parties already have access to the share of $D$ from the previous protocol, they proceed to multiply the secret shares of $D, \frac{1}{E}$, and $\Bar{I}$. Subsequently, they subtract the result from $\Bar{J}$ to obtain the final value of $(\bar{I} - \frac{D}{F} \bar{J})$.
The computation of  $(\bar{J} - \frac{D}{E} \bar{I} )$ is analogous to the one of $(\bar{I} - \frac{D}{F} \bar{J})$, where $party_2$ secretly shares $\frac{1}{E}$, and both parties participate to the multiplication of $D, \frac{1}{E}$, and $\Bar{J}$. The resulting value is finally subtracted from $\Bar{I}$, yielding the final result $(\bar{J} - \frac{D}{E} \bar{I} )$. } 
\subsection{Protocols enhancement for SSD loss}

Effectively optimizing Equation (\ref{eq: objective}) with MPC or FHE is particularly challenging, due to the computational bottleneck of these techniques when applied to large-dimensional objects \citep{haralampieva2020systematic, benaissa2021tenseal}, notably affecting the computation time and the occupation of  communication bandwidth between parties. 
% To address this issue, in what follows we introduce in the schemes of Figure \ref{fig:framework-ssd} to effectively reduce the dimensionality of image information through sampling and to improve the scalability of the algebraic operations when using these cryptographic tools.
Because cryptographic tools introduce a non-negligible overhead in terms of performance and scalability, in this section we introduce specific techniques to optimize the underlying image registration operations. 
\subsubsection{Gradient sampling}
% MI intrinsically deals with a fraction of the image content ($N_{\vx}$), while Equations (\ref{eq:jacobian_ssd}) and (\ref{eq:hessian_ssd}) are computed on the vectorized images, which are large-dimensional arrays representing all the pixels of the image (or voxels). 
Since the registration gradient is generally driven mainly by a fraction of the image content, such as the image boundaries in the case of SSD cost, a reasonable approximation of Equations (\ref{eq:jacobian_ssd}) and (\ref{eq:hessian_ssd}) can be obtained by evaluating the cost only on relevant image locations. 
This idea has been introduced in medical image registration \citep{viola1997alignment, mattes2003pet, sabuncu2004gradient}, and here is adopted to optimize Equation (\ref{eq: objective_ssd}) by reducing the dimensionality of the arrays on which encryption is performed. 
We test two different techniques: \emph{(i)}: Uniformly Random Selection (URS), proposed by \citep{viola1997alignment, mattes2003pet}, in which a random subset of dimension $l \leq d$ of spatial coordinates is sampled at every iteration with uniform probabilities, $p(\vx) = \frac{1}{d}$; and \emph{(ii)}: Gradient Magnitude Sampling (GMS) \citep{sabuncu2004gradient}, which consists of sampling a subset of coordinates with probability proportional to the norm of the image gradient, $p(\vx) \sim \| \nabla I(\vx) \|$. We note that gradient sampling is not necessary for computing the MI since in Equation (\ref{eq:joint_pdf}) the computation is already performed on a subsample of the image voxels.
\subsubsection{Matrix partitioning in FHE}
% Here we describe two protocols that can be adopted to improve the efficiency of the matrix-vector multiplication operated within FHE.
We now describe additional improvements dedicated to PPIR(FHE) and propose two versions of this solution: \emph{(i)} PPIR(FHE)-v1 implements an optimization of the matrix-vector multiplication by partitioning the vector image into a  vector of submatrices, whereas \emph{(ii)} PPIR(FHE)-v2 enhances the workload of the parties by fairly distributing the computation among them.
\paragraph{PPIR(FHE)-v1}
\label{sec:fhe_v1}
We introduce here a novel optimization dedicated to PPIR with FHE, in particular when the CKKS algorithm is adopted. CKKS allows multiple inputs to be packed into a single ciphertext to decrease the number of homomorphic operations.
To optimize matrix-vector multiplication, we propose to partition the image vector $J$ into $k$ sub-arrays of dimension $l$, and the matrix $S^T$ into $k$ sub-matrices of dimension $|\vtheta| \times l$. Once all sub-arrays $J_i$ are encrypted, we propose to iteratively apply $\textsc{DotProduct}$ as proposed by \citet{benaissa2021tenseal}, between each sub-matrix and corresponding sub-array; these intermediate results are then summed to obtain the final result, namely: $ \llbracket R \rrbracket  = \sum_{i=0}^{K} \textsc{DotProduct} \left(\llbracket J_i^T \rrbracket, S_i\right)  = S^T \cdot \llbracket J \rrbracket $.

\paragraph{PPIR(FHE)-v2}
\label{sec:fhe_v2}
% OLD VERSION: In addition to packing multiple inputs into a single ciphertext, we also propose to improve the workload among the two parties. We note that in PPIR(FHE)-v1, $party_1$ is in charge of computing the matrix-vector multiplication entirely while $party_2$ only encrypts the input and decrypts the result. In order to share the workload between both parties (Figure \ref{fig:fhe_v2}), we propose that both parties split their matrices into two sub-matrices and flatten them into vectors. $party_1$ encrypts and sends $\llbracket S_2' \rrbracket$ to $party_2$. Likewise, $party_2$ encrypts $\llbracket J_1' \rrbracket$ to $party_1$. With this new approach, both parties evaluate scalar multiplication and do not perform matrix-vector multiplication, anymore. This leads to a significant gain in terms of computational load.
In addition to packing multiple inputs into a single ciphertext, the application of FHE to PPIR can be optimized by more equally distributing the workload among the two parties. We note that in PPIR(FHE)-v1, $party_1$ is in charge of computing the matrix-vector multiplication entirely while $party_2$ only encrypts the input and decrypts the result. 
Following Supplementary Figure \ref{fig:fhe_v2}, PPIR(FHE)-v2 starts by splitting the matrix $S$ of $party_1$ into two sub-matrices $S_1$ and $S_2$ using the operation $split_{h,K}$. This operation partitions the matrix into K equally-sized sub-matrices. Next, the operation $flatten$ is applied to $S_1$ and $S_2$, obtaining vectors $S_1'$ and $S_2'$ respectively.
Then, $party_1$ encrypts $S_2'$ and sends it to $party_2$, which subsequently applies to its vector $J$ the operation $split_{v,K}$, obtaining $J_1$ and $J_2$. This operation splits $J$ into $K$ equally-sized partitions, and it executes the operation $replicate_d$ to $J_1$ and $J_2$, obtaining $J'_1$ and $J'_2$ respectively.
$party_2$ encrypts $J_1'$ and sends it to $party_1$. Both parties then iteratively perform the element-wise multiplication $\ast$ and sum up the results of the different partitions using the primitive $Sum_{i,k}$. The protocol doesn't rely anymore on \textsc{DotProduct} and leads to a significant gain in computational load.

\section{Experiments \& Results}
\label{sec:experiments}
\label{sec:implementation}

\begin{table*}[!t]
\centering
\resizebox{1.0\textwidth}{!}{ \color{black}\begin{tabular}[t]{llllllll}

\toprule
\multicolumn{8}{c}{SSD Affine registration and efficiency metrics}                                                                                                            \\ \hline
\multicolumn{1}{l|}{Solution}& \multicolumn{1}{l|}{Intensity Error (SSD)}& \multicolumn{1}{l|}{Num. Interation}& \multicolumn{1}{l|}{Displacement RMSE ($mm$)}&
\multicolumn{1}{l|}{Time  $party_1$ (s)}& \multicolumn{1}{l|}{Time $party_2$ (s)}& \multicolumn{1}{l|}{Comm. $party_1$ (MB)}& Comm. $party_2$ (MB)\\ \hline

\multicolumn{1}{l|}{\textsc{Clear}}& \multicolumn{1}{l|}{$ 4.34 \pm 0.0$} & \multicolumn{1}{l|}{$ 118 \pm 0.0 $}& \multicolumn{1}{l|}{-}& 
\multicolumn{1}{l|}{$0.0$}& \multicolumn{1}{l|}{$0.0$}& \multicolumn{1}{l|}{-}& - \\

\multicolumn{1}{l|}{\textsc{PPIR(MPC)}}& \multicolumn{1}{l|}{$ 4.34 \pm 0.0$}& \multicolumn{1}{l|}{$ 114.8 \pm 4.0$}& \multicolumn{1}{l|}{$ 0.13 \pm 0.04$}&
\multicolumn{1}{l|}{$0.13$}& \multicolumn{1}{l|}{$0.13$}& \multicolumn{1}{l|}{$ 1.54$}& $ 1.54 $ \\\hline

\multicolumn{1}{l|}{\textsc{Clear} + URS} & \multicolumn{1}{l|}{$4.38 \pm 0.0 $}& \multicolumn{1}{l|}{$61 \pm 0.0$}& \multicolumn{1}{l|}{-}&
\multicolumn{1}{l|}{$0.0 $}& \multicolumn{1}{l|}{$0.0 $}& \multicolumn{1}{l|}{-}& - \\

\multicolumn{1}{l|}{\textsc{PPIR(MPC)} + URS}& \multicolumn{1}{l|}{$ 4.34 \pm 0.0$}& \multicolumn{1}{l|}{$60.4 \pm 6.85$}& \multicolumn{1}{l|}{$ 1.75 \pm 0.19 $} &
\multicolumn{1}{l|}{$0.02$}& \multicolumn{1}{l|}{$0.02$}& \multicolumn{1}{l|}{$0.20 $}& $0.20$\\
\multicolumn{1}{l|}{\textsc{PPIR(FHE)-v1} ($D = 128$) + URS}& \multicolumn{1}{l|}{$ 4.34 \pm 0.10$}& \multicolumn{1}{l|}{$61.80 \pm 4.82$}& \multicolumn{1}{l|}{$ 13.47 \pm 2.87 $}& \multicolumn{1}{l|}{$ 2.55 $}& \multicolumn{1}{l|}{$ 0.02 $}& \multicolumn{1}{l|}{$ 0.06 $}& $ 0.01 $\\

\multicolumn{1}{l|}{\textsc{PPIR(FHE)-v2} ($D = 128$) + URS}& \multicolumn{1}{l|}{$ 4.34 \pm 0.10$}& \multicolumn{1}{l|}{$61.60 \pm 7.21$}& \multicolumn{1}{l|}{$ 13.93 \pm 4.28 $} & \multicolumn{1}{l|}{$ 0.01 $}& \multicolumn{1}{l|}{$ 0.02 $}& \multicolumn{1}{l|}{$ 0.72 $}& $ 0.88 $   \\

\hline

\multicolumn{1}{l|}{\textsc{Clear} + GMS}& \multicolumn{1}{l|}{$ 4.34 \pm 0.0 $}& \multicolumn{1}{l|}{$ 63 \pm 0.0 $} & \multicolumn{1}{l|}{-}&
\multicolumn{1}{l|}{$0.0 $}& \multicolumn{1}{l|}{$0.0 $}& \multicolumn{1}{l|}{-}& - \\ 

\multicolumn{1}{l|}{\textsc{PPIR(MPC)} + GMS}& \multicolumn{1}{l|}{$ 4.34 \pm 0.0$}& \multicolumn{1}{l|}{$59.80 \pm 6.20$}& \multicolumn{1}{l|}{$0.93 \pm 0.42$}&
\multicolumn{1}{l|}{$ 0.02$}& \multicolumn{1}{l|}{$ 0.02$}& \multicolumn{1}{l|}{$ 0.20 $}& $ 0.20 $   \\
\multicolumn{1}{l|}{\textsc{PPIR(FHE)-v1} ($D = 128$) + GMS}& \multicolumn{1}{l|}{$4.34 \pm 0.05$}& \multicolumn{1}{l|}{$60.40 \pm 5.12$}& \multicolumn{1}{l|}{$ 0.59 \pm 0.35$}& \multicolumn{1}{l|}{$ 2.51$}& \multicolumn{1}{l|}{$ 0.02$}& \multicolumn{1}{l|}{$ 0.06 $}& $ 0.01$   \\

\multicolumn{1}{l|}{\textsc{PPIR(FHE)-v2} ($D = 128$) + GMS}& \multicolumn{1}{l|}{$4.34 \pm 0.05$}& \multicolumn{1}{l|}{$57.03 \pm 4.07$}& \multicolumn{1}{l|}{$ 0.50 \pm 0.36$}& \multicolumn{1}{l|}{$ 0.02$}& \multicolumn{1}{l|}{$ 0.02$}& \multicolumn{1}{l|}{$ 0.73 $}& $ 0.93$   \\
\bottomrule

\end{tabular}}

\caption{\textcolor{black}{Affine SSD registration test, comparison between Clear, PPIR(MPC), PPIR(FHE)-v1 and PPIR(FHE)-v2. Registration metrics are reported as mean and standard deviation. Efficiency metrics in terms of average across iterations. RMSE: root mean square error.}}
\label{table:ssd-affine-results}
\end{table*}

\begin{table*}[!t]
    \centering
\resizebox{1.0\textwidth}{!}{
    \begin{tabular}[t]{llllllll}
    \toprule
    \multicolumn{8}{c}{SSD Cubic splines registration metrics} \\ \hline
    \multicolumn{1}{l|}{Solution}& \multicolumn{1}{l|}{Intensity Error (SSD)}& \multicolumn{1}{l|}{Num. Interation}& \multicolumn{1}{l|}{Displacement RMSE ($mm$)}& \multicolumn{1}{l|}{Time  $party_1$ (s)}  & \multicolumn{1}{l|}{Time $party_2$ (s)}   & \multicolumn{1}{l|}{Comm. $party_1$ (MB)} & Comm. $party_2$ (MB) \\ \hline

    \multicolumn{1}{l|}{\textsc{Clear}}& \multicolumn{1}{l|}{$ 0.65 \pm 0.0$}& \multicolumn{1}{l|}{$ 413 \pm 0.0 $}& \multicolumn{1}{l|}{-}&
    \multicolumn{1}{l|}{$0.0$}& \multicolumn{1}{l|}{$0.0 $}& \multicolumn{1}{l|}{-}& - \\
    \multicolumn{1}{l|}{PPIR-MPC}& \multicolumn{1}{l|}{$ 0.65 \pm 0.0$}& \multicolumn{1}{l|}{$ 345.70 \pm 91.22 $}& \multicolumn{1}{l|}{$ 7.31\pm 1.86$}& \multicolumn{1}{l|}{$0.63$}& \multicolumn{1}{l|}{$0.63$}& \multicolumn{1}{l|}{$ 21.47 $}& $ 28.98$ \\

        \multicolumn{1}{l|}{\textsc{PPIR(FHE)-v1}($D = 121$)}& \multicolumn{1}{l|}{$0.64 \pm 0.0$}& \multicolumn{1}{l|}{$ 224.7 \pm 79.15$}& \multicolumn{1}{l|}{$9.50 \pm 4.34 $}& \multicolumn{1}{l|}{$3.41$}& \multicolumn{1}{l|}{$ 0.00$}& \multicolumn{1}{l|}{$0.06$}& $0.01$ \\

        \multicolumn{1}{l|}{\textsc{PPIR(FHE)-v2}($D = 256$)}& \multicolumn{1}{l|}{$0.64 \pm 0.0$}& \multicolumn{1}{l|}{$ 379.2 \pm 75.82$}& \multicolumn{1}{l|}{$11.02 \pm 4.93 $}& \multicolumn{1}{l|}{$ 0.98 $}& \multicolumn{1}{l|}{$ 0.43$}& \multicolumn{1}{l|}{$ 40.45 $}& $3.56$ \\ 
    \hline

    \multicolumn{1}{l|}{Clear + URS}& \multicolumn{1}{l|}{$ 0.02 \pm 0.0 $}& \multicolumn{1}{l|}{$ 101 \pm 0.0$}& \multicolumn{1}{l|}{-}&
    \multicolumn{1}{l|}{$ 0.0$}& \multicolumn{1}{l|}{$ 0.0$}& \multicolumn{1}{l|}{-}& -\\

    \multicolumn{1}{l|}{PPIR(MPC) + URS}& \multicolumn{1}{l|}{$ 0.02 \pm 0.00$}& \multicolumn{1}{l|}{$ 79.3 \pm 1.88 $}& \multicolumn{1}{l|}{$ 5.59 \pm 0.39$}& \multicolumn{1}{l|}{$ 0.41 $}& \multicolumn{1}{l|}{$ 0.41 $}& \multicolumn{1}{l|}{$ 8.00$}& $ 8.00$\\ 
    \multicolumn{1}{l|}{\textsc{PPIR(FHE)-v1}($D = 128$) + URS}& \multicolumn{1}{l|}{$ 0.02 \pm 0.00$}& \multicolumn{1}{l|}{$ 105.40 \pm 1.71 $}  & \multicolumn{1}{l|}{$7.63 \pm 0.01 $}& \multicolumn{1}{l|}{$ 12.23 $}& \multicolumn{1}{l|}{$0.0$}& \multicolumn{1}{l|}{$ 0.06 $}& $0.01$\\ 

    \multicolumn{1}{l|}{\textsc{PPIR(FHE)-v2}($D = 128$) + URS}& \multicolumn{1}{l|}{$ 0.02 \pm 0.00$}& \multicolumn{1}{l|}{$ 105.20 \pm 2.54 $} & \multicolumn{1}{l|}{$8.74 \pm  1.90 $}& \multicolumn{1}{l|}{$ 0.62 $}& \multicolumn{1}{l|}{$0.26$}& \multicolumn{1}{l|}{$ 24.74 $}& $3.37$\\ 
    \hline

    \multicolumn{1}{l|}{Clear + GMS} & \multicolumn{1}{l|}{$ 0.02 \pm 0.0 $} & \multicolumn{1}{l|}{$ 103.00 \pm 0.0$}& \multicolumn{1}{l|}{-}& \multicolumn{1}{l|}{$ 0.0$}& \multicolumn{1}{l|}{$ 0.0$}& \multicolumn{1}{l|}{}&  \\

    \multicolumn{1}{l|}{\textsc{PPIR(MPC)} + GMS}& \multicolumn{1}{l|}{$ 0.02 \pm 0.04$}& \multicolumn{1}{l|}{$ 80.20 \pm 1.62$} & \multicolumn{1}{l|}{$  6.17 \pm 0.37$}& \multicolumn{1}{l|}{$0.41$}& \multicolumn{1}{l|}{$0.41$}& \multicolumn{1}{l|}{$8.00$}& $8.00$\\
    
    \multicolumn{1}{l|}{\textsc{PPIR(FHE)-v1}($D = 128$) + GMS}& \multicolumn{1}{l|}{$ 0.02 \pm 0.00$}& \multicolumn{1}{l|}{$ 105.70 \pm 2.40 $} & \multicolumn{1}{l|}{$5.60 \pm  2.22 $}& \multicolumn{1}{l|}{$ 11.95 $}& \multicolumn{1}{l|}{$ 0.0$}& \multicolumn{1}{l|}{$ 0.06 $}& $0.01$\\ 
    
    \multicolumn{1}{l|}{\textsc{PPIR(FHE)-v2}($D = 128$) + GMS}& \multicolumn{1}{l|}{$ 0.02 \pm 0.00$}& \multicolumn{1}{l|}{$ 106.32 \pm 1.30 $} & \multicolumn{1}{l|}{$9.11 \pm  2.34 $}& \multicolumn{1}{l|}{$ 0.62 $}& \multicolumn{1}{l|}{$ 0.26$}& \multicolumn{1}{l|}{$ 24.91 $} & $3.35$\\ 
    \bottomrule

    \end{tabular} }
    \caption{\textcolor{black}{Non-Linear SSD registration test comparison between Clear, PPIR(MPC), PPIR(FHE)-v1 and PPIR(FHE)-v2. The registration metrics are reported as mean and standard deviation. Efficiency metrics in terms of average across iterations. RMSE: root mean square error.}}
    \label{table:ssd-non-linear}
    \end{table*}

\begin{table*}[]
\color{black}
\resizebox{1.0\textwidth}{!}{

\begin{tabular}{c|c|c|c|c|c|c|c}
\toprule
\multicolumn{8}{c}{MI Affine registration metrics}                                                                                                            \\
\hline
Solution  & \multicolumn{1}{c|}{Intensity Error (MI)} & Num. Iteration & Displacement RMSE ($mm$)       & Time $party_1$ (s) & Time $party_2$ (s) & Comm. $party_1$ (MB) & Comm. $party_2$ (MB) \\
\hline
Clear     & 0.22 $\pm$ 0.00            & 213  $\pm$ 0.0          & -                            & 0.0         & 0.0          & -            & -           \\
PPIR(MPC) & 0.22 $\pm$ 0.00            & 264 $\pm$ 0.0           & 0.41 $\pm$ 0.04 & 1.02        & 1.02         & 15.00        & 15.00      \\
\bottomrule
\end{tabular}

}
\caption{\textcolor{black}{Affine MI registration test, comparison between Clear and PPIR(MPC). Registration metrics are reported as mean and standard deviation. Efficiency metrics in terms of average across iterations. RMSE: root mean square error.}}
\label{table:affine_mi}
\end{table*}
% \usepackage{multirow}
\begin{table*}[]
\color{black}
\resizebox{1.0\textwidth}{!}{

\begin{tabular}{c|c|c|c|c|c|c|c|c}
\toprule
\multicolumn{9}{c}{CC ANTs registration registration metrics}                                                                                                                                                    \\
\hline
Solution  & \multicolumn{1}{c|}{Initial DICE score}        & Final DICE score             & Num. Iteration & Displacement RMSE ($mm$)       & Time $party_1$ (s) & Time $party_2$ (s) & Comm. $party_1$ (MB) & Comm $party_2$ (MB)\\
\hline
\multicolumn{9}{c}{Forward}                                                                                                                                                                        \\
\hline
Clear     & \multirow{2}{*}{0.54 $\pm$ 0.13}                 & 0.68 $\pm$ 0.19 & 26 $\pm$ 0.0                  & -                            &  -           &     -         & -              & -            \\
PPIR(MPC) & \multicolumn{1}{l|}{}                          & 0.67 $\pm$ 0.19 & 26 $\pm$ 0.0                  & 0.22 $\pm$ 0.04 & 24.00       & 24.00        & 152.25       & 152.25      \\
\hline
\multicolumn{9}{c}{Backward}                                                                                                                                                                       \\
\hline
Clear     & \multirow{2}{*}{0.54 $\pm$ 0.13} & 0.69 $\pm$ 0.19 & 26 $\pm$ 0.0                  & -                            &   -          &   -           &     -         &   -          \\
PPIR(MPC) &                                               & 0.68 $\pm$ 0.19 & 26 $\pm$ 0.0                  & 0.22 $\pm$ 0.04 & 24.00       & 24.00        & 152.25       & 152.25 \\
\bottomrule
\end{tabular}
}
\caption{\textcolor{black}{ANTs registration with CC, comparison between Clear and PPIR(MPC). Registration metrics are reported as mean and standard deviation. Efficiency metrics in terms of average across iterations. RMSE: root mean square error.}}
\label{table:cc_ants}

\end{table*}

\color{black}

We demonstrate and assess the different versions of PPIR illustrated in Section \ref{sec:IR} on a variety of image registration problem, namely: $\emph{(i)}$ SSD for rigid transformation of point cloud data, $\emph{(ii)}$ SSD with linear and non-linear alignment of whole body positron emission tomography (PET) data; $\emph{(iii)}$ SSD and MI for mono- and multimodal linear alignment of MRI and PET brain scans; $\emph{(iv)}$ diffeomorphic non-linear registration with CC of multimodal abdomen data from CT and MRI scans. Experiments are carried out on 2D (mainly for the SSD case) and 3D imaging data. \textcolor{black}{In Table \ref{table:dataset} is reported an overview of the datasets used specifying their dimensions, modality, registration type, loss functions, and the Privacy Enhancing Technologies (PETs) employed.}

\subsection{Experimental data}
\label{subsec:dataset}
\textbf{Point Cloud Data.} 
\textcolor{black}{We showcase rigid registration on 2D point cloud data representing the corpus callosum, as presented in \citet{vachet2012automatic}, with a set size $n=193$. The registration loss here considered is SSD between point coordinates (additional details are provided in Appendix \ref{appendix:pointcloud})}.

\textbf{Whole body PET data.} The dataset considered for linear and non-linear registration with SSD consists of 18-Fluoro-Deoxy-Glucose (${}^{18}FDG$) whole body PET scans. The images are a frontal view of the maximum intensity projection reconstruction, obtained by 2D projection of the voxels with the highest intensity across views ($1260 \times 1090$ pixels).  

\textbf{Brain MRI and PET data.} This dataset regroups brain MRI and PET images obtained from the Alzheimer's Disease Neuroimaging Initiative \citep{mueller2005alzheimer}. MRI data were processed via a standard processing pipeline to estimate gray matter density maps \citep{ashburner2000voxel}. Non-linear registration was carried out on the extracted mid-coronal slice, of dimension $121 \times 121$ pixels. For 3D multimodal linear registration with MI, we use both MRI images and PET images, with respective dimension of $180 \times 256 \times 256$ and $160 \times 160 \times 96$ voxels.

\textcolor{black}{\textbf{Abdomen MR and CT data.} The  multimodal dataset Abdomen-MR-CT \citep{hering2022learn2reg} was used for experiments with ANTs registration based on CC. The data was compiled from public studies of the cancer imaging archive (TCIA) \cite{clark2013cancer} that contains 8 paired scans of MRI and CT from the same patients. The data have an isotropic resolution of $2mm$  and a voxel dimension of $192\times160\times192$.
They also provide 3D segmentation masks for the liver, spleen, and left and right kidney. All scans were pre-aligned by groupwise affine registration.}

\subsection{Experimental Details}
\label{subsec:details}
In order to avoid local minima and to decrease computation time, we use a hierarchical multiresolution optimization scheme. The scheme involves $M$ resolution steps, denoted as $r_1 \ldots r_M$. At each resolution step $r_m$, the input data is downsampled by a scaling factor $m$, where $m \in [1 \ldots M]$.
The quality of PPIR is assessed by comparing the registration results with those obtained with standard registration on clear images (\textsc{Clear}). The metrics considered are the difference in image intensity at optimum (for SSD), the total number of iterations required to converge, and the displacement root mean square difference (RMSE) between \textsc{Clear} and PPIR. We also evaluate the performance of PPIR in terms of average computation (running time) and communication (bandwidth) across iterations. \textcolor{black}{In the multimodal Abdomen data, the quality of the registration result was assessed by the overlap across the labeled anatomical regions, quantified by the DICE score}. 

\textbf{Point Cloud Data.} The registration protocol here adopted is detailed in Appendix \ref{appendix:pointcloud}. For MPC we set as the prime modulus $Q = 2^{32}$. For PPIR(FHE), we define the polynomial degree modulus as $N = 4096$. 

\textbf{Whole body PET data.}
Whole-body PET image alignment was first performed by optimizing the transformation $\mWtheta$ in Equation (\ref{eq: objective}) with respect to affine registration parameters. The multiresolution steps used are $ r_1, r_5, r_{10}, r_{20}$.
A second  whole-body PET image alignment experiment was performed by non-linear registration, without gradient approximation based on a cubic spline model (one control point every four pixels along both dimensions), with multiresolution steps $r_1$,$r_2$,$r_5$,$r_{10}$,$r_{20}$ and $r_{30}$. 
Concerning the PPIR framework, transformations were optimized for both MPC and FHE by using gradient approximation (Section \ref{sec:methods}) using the same sampling seed for each test. 
For MPC we set as the prime modulus $Q = 2^{32}$. For PPIR(FHE), we define the polynomial degree modulus as $N = 4096$, and set the resizing parameter $D$ to optimize the trade-off between run-time and bandwidth. Since $D$ needs to be a divisor of the image size image data we set $D = 128$.

\textbf{Brain MRI and PET data.}
The registration of brain gray matter density images was performed by non-linear registration based on SSD, without gradient approximation, based on a cubic spline model (one control point every five pixels along both dimensions), with multiresolution steps $ r_1 $ and $r_2$. For PPIR(MPC) we use the same configuration defined in the previous section, while for PPIR(FHE) we use the same $N$ and we set $D=121$.

We tested PPIR with MI for multi-modal 3D affine image registration between PET and MRI brain scans where, in addition to varying the multi-resolution steps, a Gaussian blurring filter is applied to the images with a kernel that narrows as multi-resolution proceeds.
The kernel size at different resolutions, denoted with $\sigma_1 \ldots \sigma_M$, is used to control the amount of blurring applied to the image at each step of the multi-resolution process. 
The multiresolution steps applied are $r_5$ and $r_{10}$, with $10\%$ of the image's pixels utilized as the number of subsample pixels ($N_{\vx}$). Gaussian image blurring is applied with a degree of $\sigma_5 = 1$ and $\sigma_{10} = 3$.
For MPC we set as the prime modulus $Q = 2^{64}$.

\textcolor{black}{\textbf{Abdomen MRI and CT data.} We tested PPIR with CC for multi-modal 3D ANTs image registration between MRI and CT abdomen scans. The multiresolution steps applied are $r_3, r_2$ and $r_1$ and the CC window size $M=5\times5\times5$. %We assess registration quality by quantifying the DICE score between the paired images both before and after the registration process.
}
\paragraph{Implementation Details} 
\textcolor{black}{The PPIR framework for the SSD is implemented using two state-of-the-art libraries:
\texttt{PySyft} \citep{ryffel2018generic}, which provides SPDZ two-party computation, and \texttt{TenSeal} \citep{benaissa2021tenseal}, which implements the CKKS protocol\footnote{\url{https://github.com/rtaiello/pp_image_registration}}}.

\textcolor{black}{PPIR based on MI and CC is implemented by extending the \texttt{Dipy} framework of \citet{garyfallidis2014dipy} \footnote{\url{https://github.com/rtaiello/pp_dipy/tree/main}}. Finally, PPIR for point cloud data is released in a separated repository\footnote{\url{https://github.com/rtaiello/ppir_pc}}.}

All the experiments are executed on a machine with an Intel(R) Core(TM) i7-7800X CPU @ (3.50GHz x 12) using 132GB of RAM. For each registration configuration, the optimization is repeated 10 times to account for the random generation of MPC shares and FHE encryption keys.

\begin{figure*}[!hbp]
\centering
  \centering
  \includegraphics[width=0.6 \linewidth]{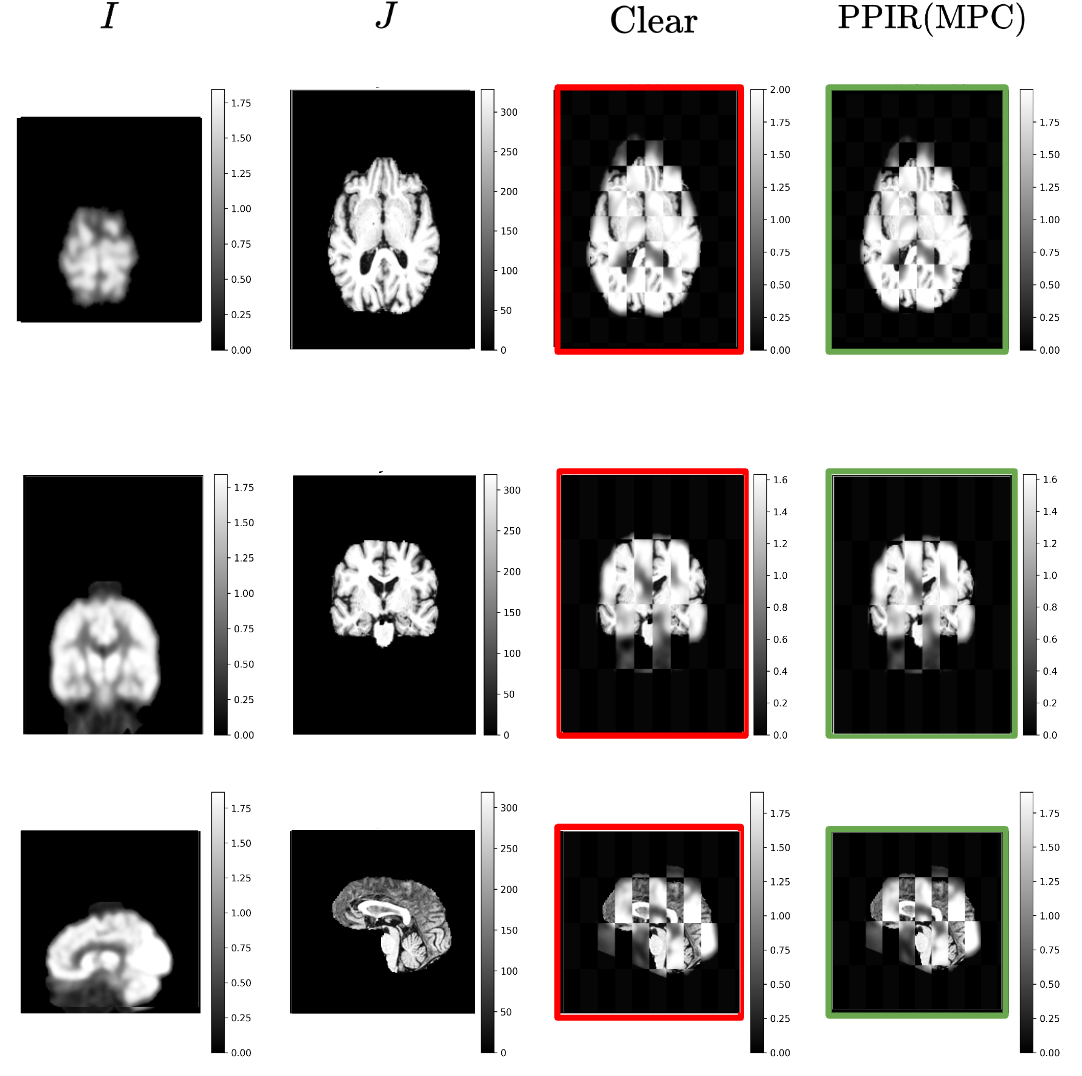}
  \caption{\textcolor{black}{Qualitative results for affine registration with MI over 3D medical images using ADNI dataset \citep{mueller2005alzheimer}. The images are presented in a $3 \times 4$ grid, with the first row representing the axial axis, the second row the coronal axis, and the third row the sagittal axis. In the first column of each row, the moving image obtained using PET modality is shown, while in the second column, the fixed image obtained using MRI modality is displayed. The third column shows the checkerboard alignment result using \textsc{Clear}, while the fourth column shows the result using \textsc{PPIR(MPC)}. The different protocols are highlighted by red and green frames, respectively.}}
  \label{fig:linear_mi}
\end{figure*}% 
\begin{figure*}[!hbp]
\centering
  \centering
  \includegraphics[width=0.7\linewidth]{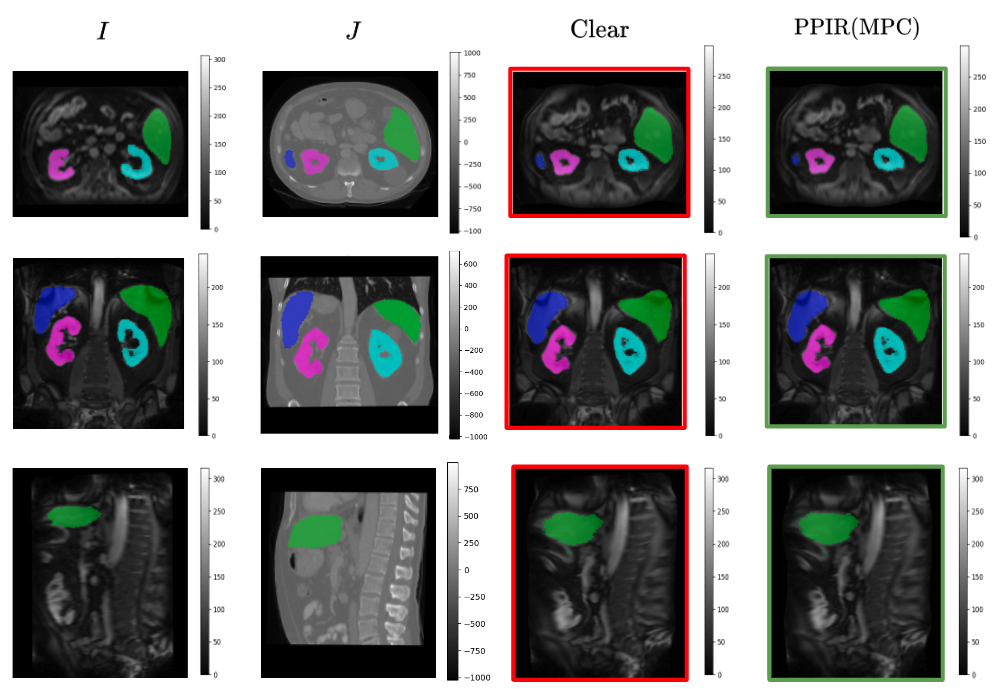}
  \caption{\textcolor{black}{Qualitative results for diffeomorphic registration with CC between 3D medical images from the AbdomenMRCT dataset \citep{hering2022learn2reg}. The images are presented in a $3 \times 4$ grid, with the first row representing the axial axis, the second row the coronal axis, and the third row the sagittal axis. First and second column show respectively MRI and CT images. The third column shows the MRI transformed using \textsc{Clear}, while the fourth column shows the MRI transformed using \textsc{PPIR(MPC)}. The transformed images are highlighted by red and green frames, respectively. }}
  \label{fig:syn_cc}
\end{figure*}% 

\subsection{Results} 
\textbf{Point Cloud Data.}  In Supplementary Table \ref{table:point_cloud} we present the registration metrics for PPIR(MPC) and PPIR(FHE)-v1. The registration shows that PPIR(MPC) achieves the best results compared to PPIR(FHE), which exhibits not only a longer computation time but also requires higher bandwidth, thanks to its non-iterative algorithm. However, to carry out \textsc{MatMul}, a sufficiently large $N$ ($4096$) is required, and in this scenario, it leads to a significant loss of chipertext slots compared to the dimension of  the point set $n=193$. Finally, the qualitative results reported in Figure \ref{fig:points_cloud} show negligible differences between point cloud transformed with \textsc{Clear}, PPIR(MPC) and PPIR(FHE)-v1.

\textbf{Whole body PET data: affine registration (SSD).} Table \ref{table:ssd-affine-results} compares \textcolor{black}{\textsc{Clear}, PPIR(MPC), PPIR(FHE)-v1 and v2}, showcasing metrics resulting from the affine transformation of whole-body PET images. Notably, registration through \textsc{PPIR(MPC)} yields negligible differences compared to \textsc{Clear} in terms of the number of iterations, intensity, and displacement.
In contrast, registering with PPIR(FHE) is not feasible when considering entire images due to computational complexity. 
Nevertheless, Supplementary Figure \ref{fig:linear_ssd} shows that neither MPC nor FHE decreases the overall quality of the affine registered images. A comprehensive assessment of the registration results is available in the Appendix. \textcolor{black}{Table \ref{table:ssd-affine-results}} (Efficiency metrics) shows that \textsc{PPIR(MPC)} performed on full images requires higher computation time and required communication bandwidth compared to PPIR(MPC)+URS/GMS. These numbers improve sensibly when using URS or GMS (by factors $10 \times$ and $20 \times$ for time and bandwidth, respectively).
Concerning \textsc{PPIR(FHE)-v1}, we note the uneven computation time and bandwidth usage between clients, due to the asymmetry of the encryption operations and communication protocol (Figure \ref{fig:framework-ssd}).
\textsc{PPIR(FHE)-v2}, which shares the computational workload between the two parties and avoids \textsc{DotProduct}, allows obtaining an important speed-up over \textsc{PPIR(FHE)-v1} ($100\times$ faster). Notably, this gain is obtained without affecting the quality of the registration metrics, and improves the execution time of \textsc{PPIR(MPC)}. On the other side, although \textsc{PPIR(FHE)-v1} is able to improve communication with respect to \textsc{PPIR(MPC)}, it still suffers from the highest \textcolor{black}{communication} among the three proposed solutions. This is due to the fact PPIR(FHE) protocols can find different applications depending on the requirements in terms of computational power or bandwidth.

\textbf{Brain MRI data and whole body PET data: non-linear registration (SSD).}
\textcolor{black}{Table \ref{table:ssd-non-linear}, comparing \textsc{Clear} and \textcolor{black}{PPIR(MPC), PPIR(FHE)-v1 and v2}, showcases the metrics resulting from spline-based non-linear registration between grey matter density images without the application of gradient approximation. Additionally, the table includes results for the registration between whole-body PET images when the gradient approximation is applied.}

\textit{Brain MRI data without gradient approximation.}
\textcolor{black}{
Regarding the registration accuracy, we draw conclusions similar to those of the affine case, where \textsc{PPIR(MPC)} leads to minimum differences with respect to \textsc{Clear}, while \textsc{PPIR(FHE)-v1} seems slightly superior. 
\textsc{PPIR(MPC)} is associated with a lower execution time and a higher computational bandwidth, due to the larger number of parameters of the cubic splines, which affects the size of the matrix $S$. 
Although \textsc{PPIR(FHE)-v1} has a slower execution time, the demanded bandwidth is inferior to the one of \textsc{PPIR(MPC)}, since the encrypted image is transmitted only once.
\textsc{PPIR(FHE)-v2}, as in the affine case, outperforms \textsc{PPIR(FHE)-v1} (still about $100\times$ faster) leading to comparable values between registration metrics and is still inferior to \textsc{PPIR(MPC)}. 
Here, the limitations of \textsc{PPIR(FHE)-v1} on the bandwidth size are even more evident than in the affine case, since the bandwidth increases according to the number of parameters. This result gives a non-negligible burden to the $party_1$, due to the multiple sending of the flattened and encrypted submatrices of updated parameters. Furthermore, in this case, \textsc{PPIR(FHE)-v1} performs slightly worse than \textsc{PPIR(MPC)} in terms of execution time. }

\textit{Whole Body PET Data With Gradient Approximation.}
\textcolor{black}{
Incorporating gradient approximation for handling whole-body PET data leads to similar conclusions as for the experiments on brain data. Qualitative results, reported in Supplementary Figure \ref{fig:non_linear_ssd}, show negligible differences between images transformed with \textsc{Clear}+GMS, \textsc{PPIR(MPC)}+GMS, and \textsc{PPIR(FHE)-v1}+GMS.}

\textbf{Brain MRI and PET data: affine registration (MI).}
\textcolor{black}{Table \ref{table:affine_mi} provides information on the computation cost of the protocol and the registration metrics for both the joint PDF and the First Derivative of the Joint PDF, using both \textsc{Clear} and \textsc{\textsc{PPIR(MPC)}}. The results demonstrate a reasonable execution time (completed in under 5 minutes for the entire process) and noteworthy data transfer, totaling less than $4GB$.
Qualitative results for the image registration are shown in Figure \ref{fig:linear_mi}, indicating that there is no difference between the moving transformed using \textsc{Clear} and \textsc{PPIR(MPC)}.}

\textcolor{black}{
\textbf{Abdomen MRI and CT data: diffeomorphic non-linear registration (CC).}
Table \ref{table:cc_ants} presents the metrics for ANTs registration using both \textsc{Clear} and \textsc{PPIR(MPC)} methods. The initial DICE scores for both forward and backward transformations are consistent between \textsc{Clear} and \textsc{PPIR(MPC)}, with slight variations in the final DICE scores. The number of iterations and displacement RMSE values also exhibit similar trends.
In terms of communication and computation times, \textsc{PPIR(MPC)} demonstrates comparable performance with \textsc{Clear}, showcasing its feasibility for secure registration. The computation times and communication bandwith between the two parties are well within reasonable limits.
These qualitative result in Figure \ref{fig:syn_cc} indicates that the proposed \textsc{PPIR(MPC)} solution maintains the quality of registration metrics while ensuring secure and private communication between parties.}

\section{Discussion}
\textcolor{black}{
This current work builds upon \citet{10.1007/978-3-031-16446-0_13} by extending its application to include MI with linear registration (3D images), CC using the ANTs framework (3D images), enhancing the FHE for the SSD loss function, namely PPIR(FHE)-v2, and also integrating rigid point cloud registration.
We recognize specific limitations associated with the use of FHE in our PPIR framework, which limited the effective use of this techniques besides the optimization of the SSD loss. FHE's computational cost during homomorphic operations poses challenges, limiting the scalability and real-time applicability of PPIR(FHE). This is particularly evident when dealing with large datasets, such as 3D images, or when employing advanced image registration cost functions that demand significant computational resources. To address this gap, researchers are actively exploring the optimization of FHE through the integration of hardware accelerators \citep{boemer2021intel}.}

\textcolor{black}{
For the SSD loss function, we provide comparison experiments with both URS and GMS \citep{viola1997alignment, mattes2003pet, sabuncu2004gradient} for sake of completeness and compatibility with subsampling approaches in IR. We recognized that URS doesn’t bring substantial improvements with resepct to GRS, and this latter method should be preferred in the considered application or testing scenario.}

\textcolor{black}{
While PPIR focuses on the privacy-preserving formulation of classical image registration methods based on gradient-based optimization, throughout the past years the research community has been steering the attention towards deep learning (DL)-based image registration  \citep{simonovsky2016deep, krebs2017robust, yang2017quicksilver, balakrishnan2019voxelmorph}. 
% While a number of privacy-preserving DL training approaches have been proposed in the medical imaging literature, DL-based registration frameworks are currently missing, and represent a complementary research direction to the PPIR methodology proposed in this work, which comes with specific challenges.  
Among the medical imaging application of privacy-preserving methodologies,  %\citet{terrail2022flamby} proposed a collaborative model training using Federated Learning (FL) for a 3D medical image segmentation task. Additionally, 
\citet{kaissis2021end} discussed privacy-preserving FL with Secure Aggregation \citep{bonawitz2017practical} and Differential Privacy \citep{abadi2016deep} for 2D medical image classification tasks. However, as highlighted by \citep{kaissis2021end}, deploying DL models for privacy-preserving inference nowadays is predominantly achievable through Multi-Party Computation (MPC). This process necessitates multiple servers and incurs significant overhead, primarily attributed to the size of the DL model, especially when handling 3D image registration tasks within a DL-based framework \citep{balakrishnan2019voxelmorph}.
\textcolor{black}{To the best of our knowledge both DL and non-DL registration methods available in the literature do not satisfy the PPIR requirements investigated in our work, as they always require the disclosure of the target and moving images in clear.}
\section{Conclusion and future works} 
This study introduces the novel paradigm of Privacy Preserving Image Registration, designed for allowing image registration in privacy-preserving scenarios where images are confidential and cannot be shared in clear. Leveraging both secure multi-party computation (MPC) and Fully Homomorphic Encryption (FHE), we propose in PPIR effective strategies integrating cryptographic techniques into a variety of state-of-the-art registration frameworks, encompassing different parameterization and loss functions. We evaluate the framework's performance across various registration benchmarks, conducting quantitative and qualitative assessment for all the considered image registration problems.
Our future direction involve extending PPIR to encompass additional cost functions commonly used in image registration, aiming to enhance the framework's versatility and applicability.}

\section*{Acknowledgments}
This work has been supported by the French government, through the 3IA Côte d’Azur Investments in the Future project managed by the National Research Agency (ANR) with the reference number ANR-19-P3IA-0002, and by the ANR JCJC project Fed-BioMed 19-CE45-0006-01. 
%%Harvard
\bibliographystyle{model2-names.bst}\biboptions{authoryear}
\bibliography{biblio}

\pagebreak
\begin{center}
\textbf{\large Appendix}
\end{center}
%%%%%%%%%% Merge with supplemental materials %%%%%%%%%%
%%%%%%%%%% Prefix a "S" to all equations, figures, tables and reset the counter %%%%%%%%%%
\setcounter{equation}{0}
\setcounter{figure}{0}
\setcounter{table}{0}
\setcounter{page}{1}
\setcounter{section}{0}
\makeatletter
\renewcommand{\theequation}{A\arabic{equation}}
\renewcommand{\thefigure}{A\arabic{figure}}
\renewcommand{\thetable}{A\arabic{table}}
\renewcommand{\bibnumfmt}[1]{[A#1]}
\renewcommand{\citenumfont}[1]{A#1}
\renewcommand{\thesection}{A~\arabic{section}}
\renewcommand{\thesubsection}{\arabic{section}.\arabic{subsection}}

% \begin{figure*}[!htp]
% \centering

% \begin{subfigure}{.5\textwidth}
%   \centering
%   \includegraphics[width=.67\linewidth]{joint_deriv_pdf_fhe.drawio.png}
%   \caption{PPIR(FHE)}
%   \label{fig:joint_deriv_pdf_fhe}
% \end{subfigure}
% \caption{Optimization of MI loss: proposed framework to calculate matrix multiplication $P'= - \frac{1}{N_{\vx}} \cdot (B_J^0)^T \cdot C_I^3$ based on PPIR(MPC) and PPIR(FHE).}
% \label{fig:ppir_joint_deriv_pdf}
% \end{figure*}
\begin{figure*}[!htp]
\centering
\begin{subfigure}{.5\textwidth}
  \centering
  \includegraphics[width=.65\linewidth]{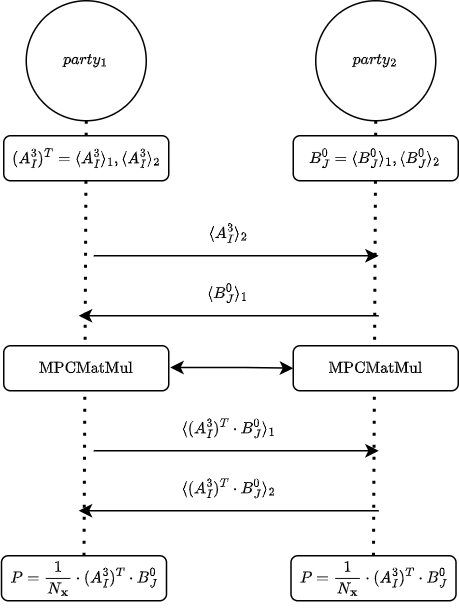}
  \caption{Computation $ P = \frac{1}{N_{\vx}} \cdot (A_I^3)^T \cdot B_J^0$ with PPIR(MPC) }
  \label{fig:joint_pdf_mpc}
\end{subfigure}%
\begin{subfigure}{.5\textwidth}
  \centering
  \includegraphics[width=.65\linewidth]{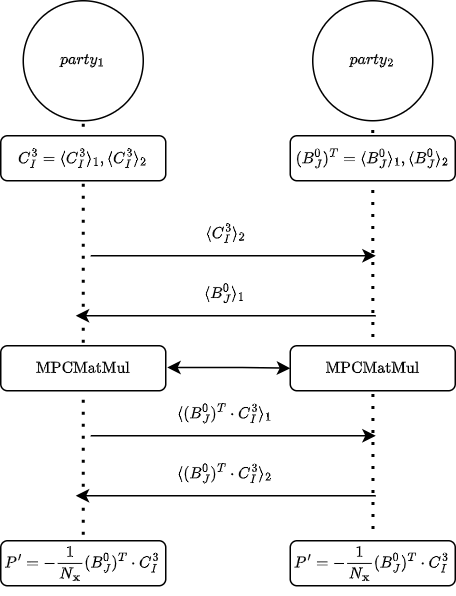}
  \caption{Computation $P'= - \frac{1}{N_{\vx}} \cdot (B_J^0)^T \cdot C_I^3$ with PPIR(MPC)}
  \label{fig:joint_deriv_pdf_mpc}
\end{subfigure}%
\caption{Optimization of MI loss: proposed framework to calculate matrix multiplication $ P = \frac{1}{N_{\vx}} \cdot (A_I^3)^T \cdot B_J^0$ and $P'= - \frac{1}{N_{\vx}} \cdot (B_J^0)^T \cdot C_I^3$ based on PPIR(MPC).}
\label{fig:ppir_joint_pdf}
\end{figure*}
\begin{figure*}[!htp]
\centering
\begin{subfigure}{.5\textwidth}
\vskip 0pt
  \centering
  \includegraphics[width=.5\linewidth]{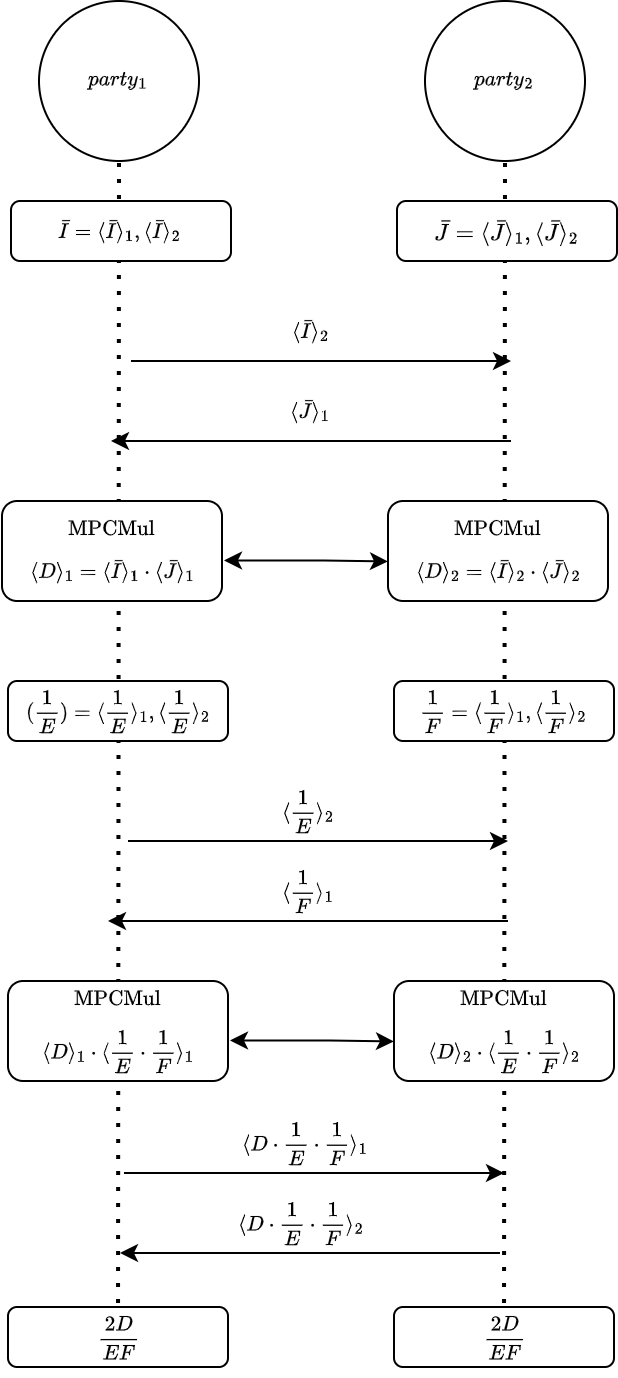}
  \caption{PPIR(MPC) - $\frac{2D}{EF}$ }
\label{fig:joint_deriv_cc}
\end{subfigure}%
\begin{subfigure}{.5\textwidth}
\vskip 0pt
  \centering
  \includegraphics[width=.57\linewidth]{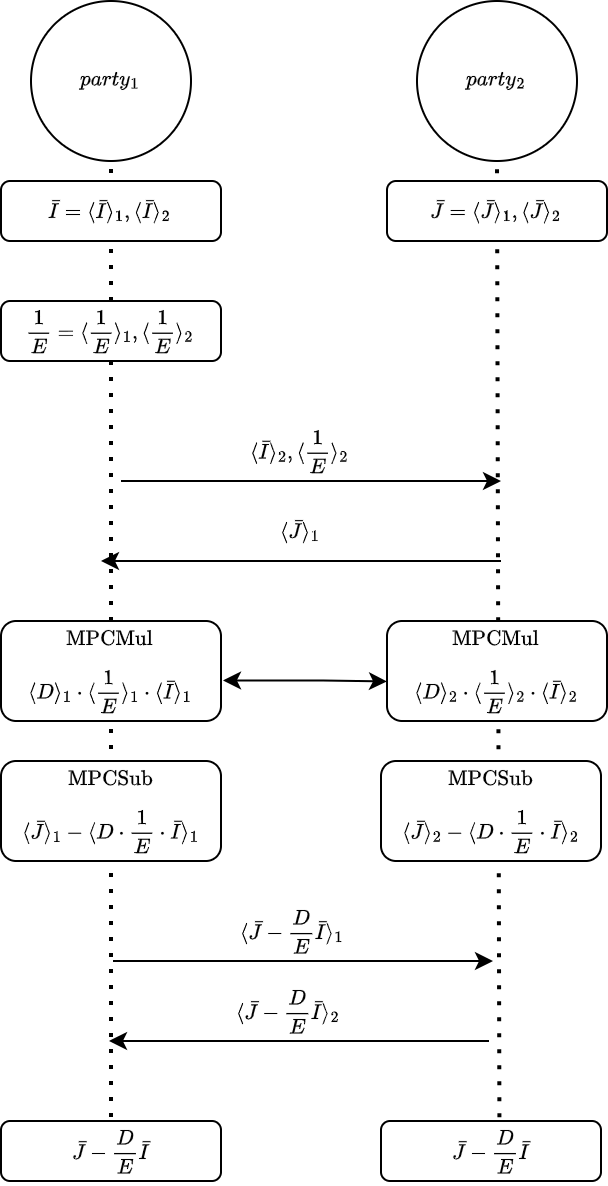}
  \caption{PPIR(MPC) - $(\bar{J} - \frac{D}{E} \bar{I} )$}
 \label{fig:fw_cc}
\end{subfigure}
\caption{\textcolor{black}{Optimization of ANTS NCC loss: proposed framework to calculate $\frac{2D}{EF}$ and $(\bar{J} - \frac{D}{E} \bar{I} )$ based on PPIR(MPC).}}
\label{fig:ppir_fw_bw_cc}
\end{figure*}

\begin{figure*}[!htp]
\centering
\begin{subfigure}{.5\textwidth}
  \centering
  \includegraphics[width=.65\linewidth]{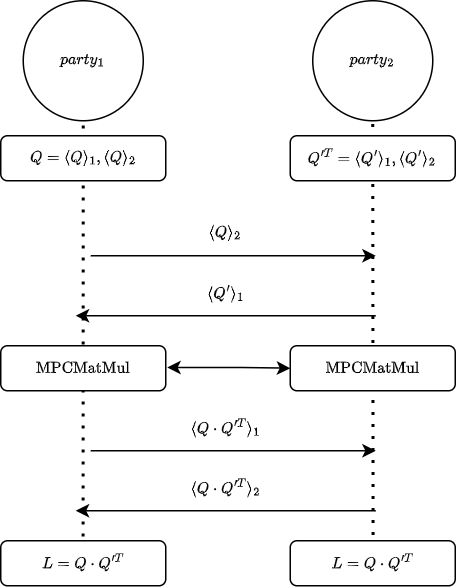}
  \caption{PPIR(MPC)}
\end{subfigure}%
\begin{subfigure}{.5\textwidth}
  \centering
  \includegraphics[width=.67\linewidth]{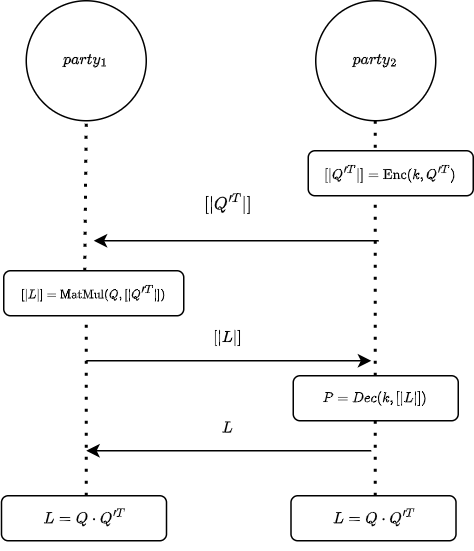}
  \caption{PPIR(FHE)}
\end{subfigure}
\caption{\textcolor{black}{Optimization of rigid point cloud: proposed framework to compute matrix multiplication $L = Q \cdot Q'^T$ based on PPIR(MPC) and PPIR(FHE).}}
\end{figure*}

\begin{table}[]
\color{black}
    \resizebox{0.5\textwidth}{!}{
\begin{tabular}{c|c|c|c|c|c}
\toprule
\multicolumn{6}{c}{Rigid Point Clouds Registration}                                                                                                                                                \\
\hline
Solution    & \multicolumn{1}{c|}{Displacement RMSE ($mm$)} & \multicolumn{1}{c|}{Time $party_1$ ($s$)} & \multicolumn{1}{c|}{Time $party_2$ ($s$)} & \multicolumn{1}{c|}{Comm. $party_1$ (MB)} & \multicolumn{1}{c}{Comm. $party_2$ (MB)} \\
\hline
PPIR(MPC)   & 1.11 $\pm$ 0.31               & 0.02                            & 0.02                             & 0.03                             & 0.03                            \\
PPIR(FHE) & 1.26 $\pm$ 0.37               & 1.10                            & 0.18                             & 1.42                             & 35.93 \\
\bottomrule
\end{tabular}
}
\caption{\textcolor{black}{Rigid Point Clouds registration test, comparisson between PPIR(MPC) and PPIR(FHE). Registration metrics are reported as mean and standard deviation. Efficiency metrics in terms of average across iterations. RMSE: root mean square error.}}
\label{table:point_cloud}

\end{table}
\color{black}

\begin{figure*}
\centering
  \centering
  \includegraphics[width=0.9\linewidth]{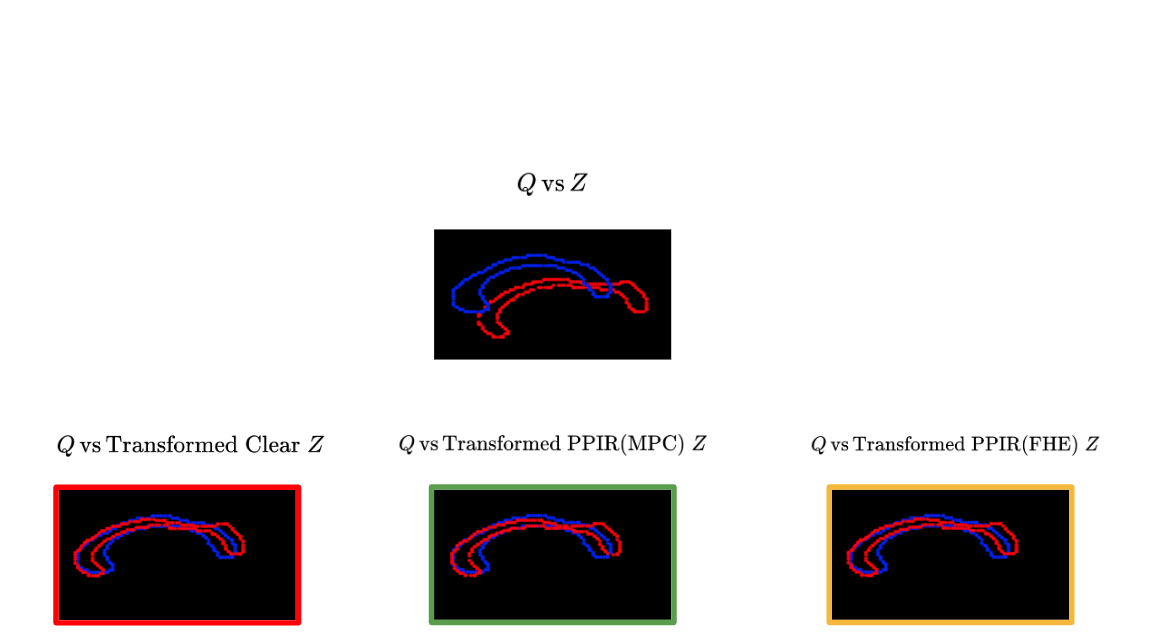}
  \caption{Qualitative results for rigid point cloud registration between 2D corpus callosum point sets. The red frame is the transformed moving image using \textsc{Clear} registration. Green and Yellow frames are the transformed images using respectively PPIR(MPC)  and PPIR(FHE)v1.}
  \label{fig:points_cloud}
\end{figure*}

\textcolor{black}{
\section{Rigid Point Cloud Registration}
\label{appendix:pointcloud}
Let $\{z_i\}$, $\{z'_i\}$ two finite $n$ size point sets where $ z_i, z'_i \in \mathbb {R} ^{d}$ , to continue ...
A non-iterative least-squares approach to match two sets of points, was proposed by \citet{arun1987least}.
The method uses singular value decomposition (SVD) and is trying to minimize the following cost function:
\begin{equation}
    \Sigma^{2} = \sum_{i=1}^{n} || z'_{i} - (R z_{i} + \vt) ||^{2},
\end{equation}
where $R$ is the rotation matrix and $\vt$ is the translation vector.
Let define $\bar{z}$ and $\bar{z'}$ to represent the centroids of $\{z_i\}$ and $\{z'_i\}$ respectively.
Let $\hat{R}$ being the estimated rotation matrix, and $\hat{\vt}$ being the estimated translation.
The method lies in the algorithm for finding $\hat{R}$ detailed below:
\begin{enumerate}
    \item Calculate the following quantities:
    \begin{align*}
        q_i = p_i - \bar{z} \\
        q'_i = p'_i - \bar{z'} 
    \end{align*}
    for all $0\leq i \leq n$, which are distances from each point to its centroid.
    \item Calculate $L\in \mathbb{R}^{d \times d}$, $L = \sum_{i=1}^{n} q_{i} q_{i}'^{T}$, in vectorized form:
        \begin{equation}
            L = Q \cdot Q'^T, 
            \label{eq:point_cloud}
    \end{equation}
    where $Q, Q' \in \mathbb{R}^{d \times n}$.
    \item Find SVD of $L$, namely $L = U \Lambda V^{T}$;
    \item Calculate $X = VU^T$;
    \item Check the determinant of $X$. If it equals to $+1$, then $\hat{R} = X$ and $\hat{\vt}  = q' - \hat{R} q$. If the determinant equals to $-1$, the algorithm has failed.
\end{enumerate}
We note that the only operation that requires the joint availability of information from both parties is Equation \ref{eq:point_cloud} which can be computed with a matrix multiplication with PPIR(MPC) and PPIR(FHE) as reported in Figure \ref{fig:points_cloud}.}
%{\paragraph{Experiments \& Results} %For the rigid point cloud registration we used 2D point cloud representing corpus callosum defined in \citet{vachet2012automatic}, with a set size $n=193$.
%In Table \ref{table:point_cloud} we presesent the registration metrics for PPIR(MPC) and PPIR(FHE)-v1}. The registration shows that PPIR(MPC) achieves the best results compared to PPIR(FHE), which exhibits not only a longer computation time but also boasts higher bandwidth, thanks to its non-iterative algorithm. However, to carry out \textsc{MatMul}, a sufficiently large $N$ ($4096$) is required, and in this scenario, it leads to a significant loss of chipertext slots compared to the dimension of  the point set $n=193$. Finally, the qualitative results, reported in Figure \ref{fig:points_cloud}, shows negligible differences between point cloud transformed with \textsc{Clear}, PPIR(MPC) and PPIR(FHE)-v1.}
\begin{figure*}
  \centering
  \includegraphics[width=.70\linewidth]{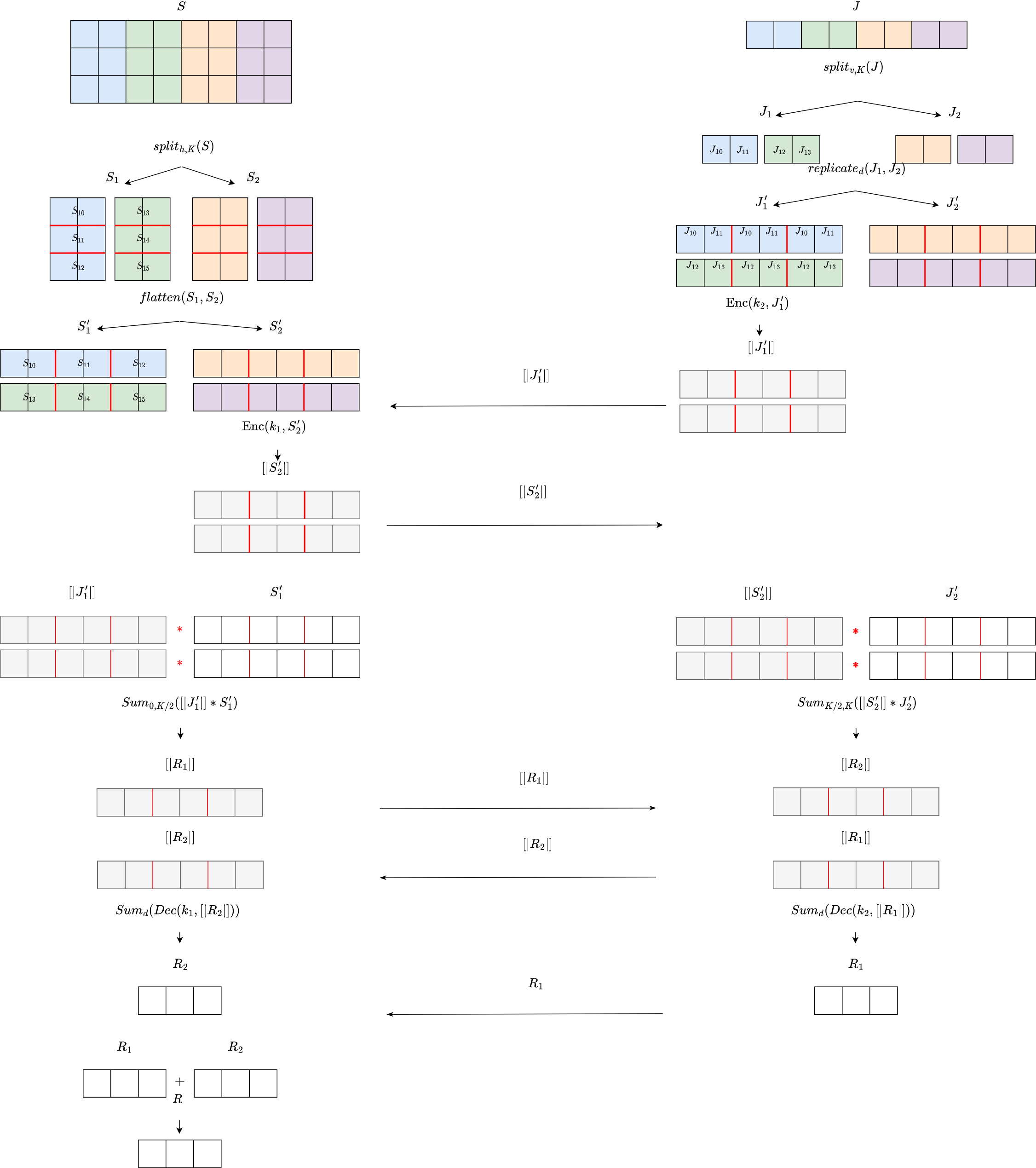}
  \caption{Proposed framework to compute matrix-vector multiplication $S^T \cdot J$ based on PPIR(FHE)-v2.}
  \label{fig:fhe_v2}
\end{figure*}%

\begin{figure*}
\centering
  \centering
  \includegraphics[width=.8\linewidth]{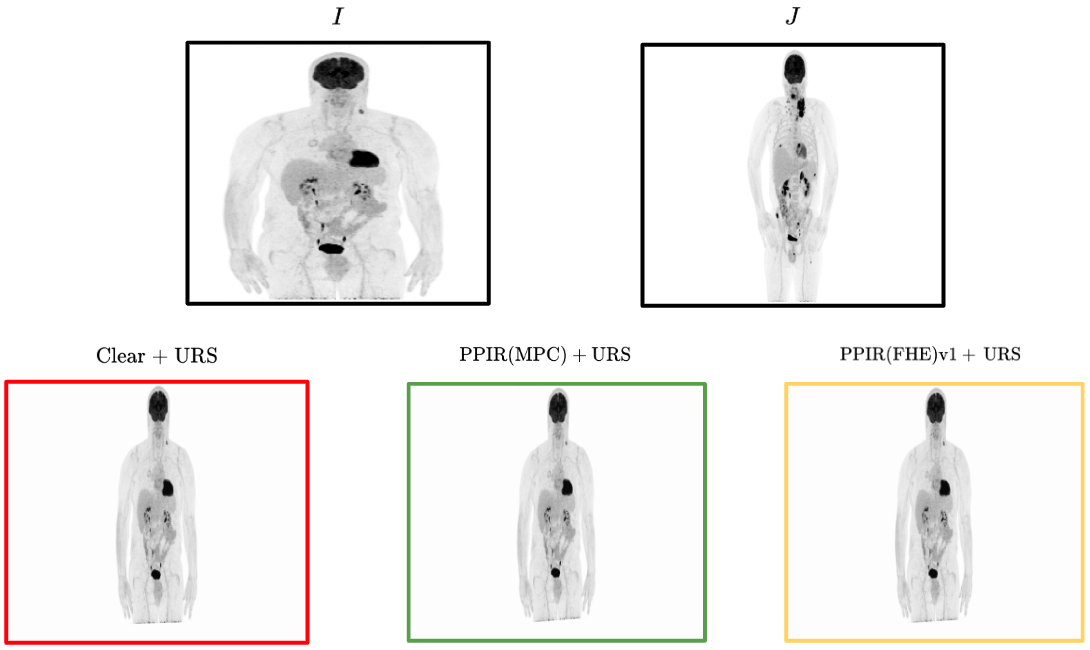}
  \caption{Qualitative results for affine registration with SSD between 2D medical images. The red frame is the transformed moving image using \textsc{Clear}+URS registration. Green and Yellow frames are the transformed images using respectively PPIR(MPC)+URS  and PPIR(FHE)v1+URS.}
  \label{fig:linear_ssd}
\end{figure*}%

\begin{figure*}
\centering
  \centering
  \includegraphics[width=.9\linewidth]{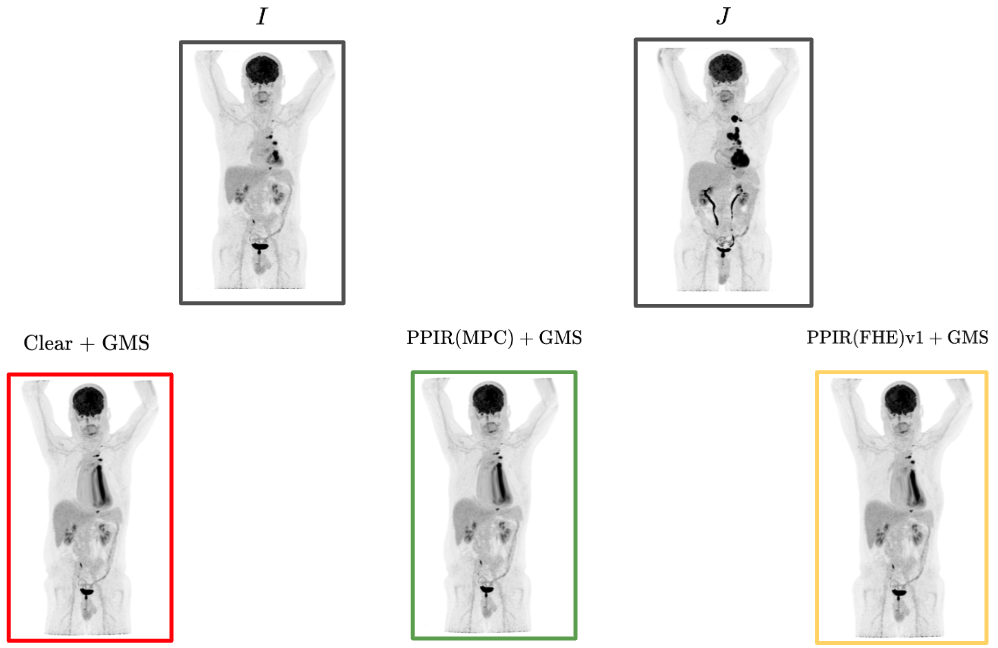}
  \caption{Qualitative results for Cubic splines registration with SSD between 2D medical images. The red frame is the transformed moving image using \textsc{Clear}+GMS registration. Green and Yellow frames are the transformed images using respectively PPIR(MPC)+GMS  and PPIR(FHE)v1+GMS.}
  \label{fig:non_linear_ssd}
\end{figure*}% 

\end{document}